%% file: preprint.tex
\documentclass{article}
\usepackage{iclr2027_conference,times}

\iclrfinalcopy

\input{math_commands.tex}

\usepackage{url}

\usepackage{amsmath,amssymb,amsfonts}
\usepackage{booktabs}
\usepackage{multirow}
\usepackage{graphicx}
\usepackage{xspace}
\usepackage{xcolor}
\usepackage[colorlinks=true,linkcolor=blue!80!green,citecolor=blue!80!green,urlcolor=blue!80!green]{hyperref}

\newcommand{\teacher}{\textsc{TID}\xspace}
\newcommand{\student}{\textsc{TIDE}\xspace}

\newcommand{\teacherfull}{\emph{\textbf{T}raining-data \textbf{I}nfluence via score \textbf{D}iscrepancy}\xspace}
\newcommand{\studentfull}{\emph{\textbf{TID} \textbf{E}mbeddings}\xspace}

\usepackage{amsthm}
\usepackage{dsfont}
\usepackage{algorithm}
\usepackage{algpseudocode}
\newtheorem{lemma}{Lemma}

\usepackage{anyfontsize}
\renewcommand{\paragraph}[1]{\vspace{0.7mm}\textbf{#1} }

\title{Distilling Diffusion Score Discrepancy for Efficient Training Data Attribution}

\author{
\textbf{Shixuan Liu}$^{1,2}$\thanks{Work done during an internship at Sony Group Corporation.},
\textbf{Joan Serr{\`a}}$^{3}$,
\textbf{Kin Wai Cheuk}$^{2}$,
\textbf{Jinju Kim}$^{2,4}$\footnotemark[1],
\textbf{Woosung Choi}$^{2}$\\[0.6em]
\textbf{Yukara Ikemiya}$^{2}$,
\textbf{Wei-Hsiang Liao}$^{2}$,
\textbf{Jiaqi W. Ma}$^{1}$,
\textbf{Yuki Mitsufuji}$^{5}$\\[1em]
{\normalfont $^{1}$University of Illinois at Urbana-Champaign}\\
{\small\normalfont \texttt{\{shixuanl,jiaqima\}@illinois.edu}}\\[0.5em]
{\normalfont $^{2}$Sony Group Corporation \quad $^{3}$Sony Europe \quad $^{5}$Sony CTC America}\\
{\small\normalfont \texttt{\{joan.serra,kinwai.cheuk,woosung.choi,}}\\
{\small\normalfont \texttt{yukara.ikemiya,weihsiang.liao,yuki.mitsufuji\}@sony.com}}\\[0.5em]
{\normalfont $^{4}$University of Texas at Austin}\\
{\small\normalfont \texttt{jinju.is.pearl@utexas.edu}}
}

\begin{document}

\maketitle
\lhead{Preprint}

\begin{abstract}
Training data attribution for diffusion models aims to identify the training samples that influence a generated instance, but existing methods either require costly per-sample gradient computation or query-specific model optimization. Moreover, most methods attribute changes in a proxy loss rather than changes in the actual model's generative behavior. We address these limitations by formulating attribution directly with a \emph{local score discrepancy} measure, which applies to any diffusion variant (including DDPM, EDM, and flow matching), and by showing that such measure can be estimated without retraining, as a preconditioned gradient similarity. We instantiate this estimator as \teacherfull (\teacher), which uses Kronecker-factored curvature to avoid random projections and per-sample gradient storage. We then distill \teacher{} into \student{}, a forward-only student trained online to reproduce the teacher's rankings from the diffusion model's internal activations. Under counterfactual evaluation on CIFAR-10, ArtBench-10, and MS-COCO, \teacher{} matches or outperforms state-of-the-art approaches, while \student{} retains most of \teacher{}'s accuracy at four to five orders of magnitude lower per-query cost, attributing generated samples in milliseconds and faster than the generation itself. 

\end{abstract}

\input{sections/intro}

\input{sections/method}

\input{sections/experiments}

\input{sections/conclusion}

\subsection*{AI use statement}

In this work, we have not used generative AI tools to generate synthetic data sets, help develop theoretical models or conceptual frameworks, formulate mathematical claims, provide critical ingredients for proving mathematical claims, assist in the writing of proofs, propose or refine hypotheses, design or provide feedback on research methodology or experiments, interpret results, assist with translation, clean and reformat datasets, or support qualitative and thematic data analysis. We have used generative AI tools to create or edit code, draft parts of the research paper, and edit the research paper to improve readability. We have reviewed all AI-assisted work: LLM-polished and LLM-drafted content was proofread and further improved by the authors, and LLM-generated code was verified and tested for correctness by the authors. We take responsibility for the final content of this work, including text, claims, or artifacts produced with the aid of generative AI.

\subsection*{Ethics statement}

This work does not involve human-subject studies, the collection of personal or sensitive data, or the creation or release of new datasets. It does not conduct interventions involving individuals or deploy systems to make decisions about them. Our methods aim to support transparency by estimating the influence of training examples on generated outputs. Practitioners should exercise caution when using these estimates to allocate compensation to data contributors or assess potential copyright infringement. Attribution scores alone do not establish copying or entitlement to compensation and should be considered alongside independent evidence and the relevant context.

\subsection*{Reproducibility statement}

All experiments use fixed random seeds, and results are reported over multiple seeds. Full experimental details are given in Appendix~\ref{app:exp-details}, and the derivations supporting our method are given in Appendix~\ref{app:method-proof}. We will release the code upon acceptance of the paper.

\bibliography{iclr2027_conference}
\bibliographystyle{iclr2027_conference}

\newpage
\appendix
\input{sections/appendix}

\end{document}

%% file: math_commands.tex
\usepackage{amsmath,amsfonts,bm}

\def\eqref#1{equation~\ref{#1}}

\def\1{\bm{1}}

\DeclareMathAlphabet{\mathsfit}{\encodingdefault}{\sfdefault}{m}{sl}
\SetMathAlphabet{\mathsfit}{bold}{\encodingdefault}{\sfdefault}{bx}{n}

%% file: sections/intro.tex
\section{Introduction}
\label{sec:intro}

Diffusion models are trained on large-scale collections of human-created images and have demonstrated remarkable capabilities in synthesizing diverse, high-quality visual content~\citep{ho2020denoising,song2020denoising,karras2022elucidating,lipman2022flow}. Their widespread adoption, however, has also raised growing concerns regarding copyright infringement~\citep{zhang2023copyright,lu2024disguised}, memorization~\citep{somepalli2023diffusion,carlini2023extracting}, and other data-related issues. These concerns have motivated increasing interest in training data attribution (TDA) for diffusion models~\citep{dai2023training,georgiev2023journey,zheng2024intriguing,wang2024data,lin2025diffusion,serra2026training}, which aims to identify the training samples that most strongly influence a particular generated image.

Despite the growing body of work on diffusion TDA, two main issues remain open. The first concerns the formulation: most gradient-based estimators measure the counterfactual effect of a training sample on a proxy utility, typically the training loss~\citep{georgiev2023journey,zheng2024intriguing}, whereas the question asked of a generative model is how its training data determines what it generates. A shift in loss does not necessarily reflect a shift in the generated output~\citep{lin2025diffusion}. The second concerns the cost: gradient-based attribution~\citep{georgiev2023journey,zheng2024intriguing,lin2025diffusion} requires per-sample gradients for both training and query samples across multiple diffusion timesteps, at a cost that can approach that of model training. Unlearning-based attribution~\citep{ko2024mirrored,wang2024data,serra2026training} reduces the training side to forward passes only, but requires a dedicated fine-tuning procedure for each query, so the overall cost still does not scale. Appendix~\ref{sec:related-work} provides a detailed review of existing approaches.

In this work, we address both aforementioned issues. For the formulation issue, we start from the observation that sampling in a diffusion model is driven by its learned score~\citep{song2020score,song2020improved}, so removing a training sample can change what is generated for a query mainly by perturbing the score along the query's forward noising. We therefore define the influence of a training sample as the expected \emph{score discrepancy} between the original and the perturbed model \emph{at different noisy states of the query}. This target is generalizable for most diffusion variants, including DDPM~\citep{ho2020denoising}, EDM~\citep{karras2022elucidating}, and flow matching~\citep{lipman2022flow,liu2022flow,albergo2022stochastic}, and a first-order expansion turns it into a preconditioned gradient similarity that can be evaluated without retraining. We instantiate this estimator as \teacherfull (\teacher). To avoid the distortion introduced by random projections, \teacher preconditions gradients with layer-wise Kronecker-factored curvature~\citep{martens2015optimizing,george2018fast}, which also removes the need for per-sample gradient storage.

For the cost issue, we follow the recent idea of influence distillation~\citep{wang2026fast} and distill \teacher{} into a forward-only student model built directly from the diffusion model's internal representations, eliminating the need for a separate visual encoder. We term this model \studentfull (\student). Rather than constructing a fixed distillation dataset offline~\citep{wang2026fast}, our TID formulation enables the generation of teacher supervision online within each training batch. This strategy results in efficient post hoc training while exposing the student to a broader and continually evolving set of attribution targets. TIDE produces compact, low-dimensional embeddings that support efficient nearest-neighbor retrieval under standard similarity metrics, the core operation required to attribute a generated instance.\footnote{Our code will be available soon.}

Our main contributions are as follows:
\begin{itemize}
\item We formulate diffusion TDA as the expected local score discrepancy induced by removing a training sample from a query's forward noising, a target that is expressed directly in the model's generative mechanism and applies across multiple diffusion variants.
\item We instantiate this target as \teacher (Sec.~\ref{sec:method-teacher}), a gradient-based teacher that uses Kronecker-factored curvature to avoid both random projections and per-sample gradient storage.
\item We distill \teacher into \student (Sec.~\ref{sec:method-student}), a forward-only student trained online from the diffusion model's own activations using a top-heavy ranking objective.
\item Counterfactual retraining on CIFAR-10, ArtBench-10, and MS-COCO shows that \teacher outperforms state-of-the-art unlearning-based TDA on two datasets, while \student retains most of its accuracy at four-to-five orders of magnitude lower per-query cost. Ablations validate our core design choices (Secs.~\ref{sec:res-existing}--\ref{sec:res-ablation}).
\end{itemize}

%% file: sections/method.tex
\section{Method}
\label{sec:method}

In this section, we introduce \teacherfull{} (\teacher{}, Sec.~\ref{sec:method-teacher}), a general gradient-based attribution method applicable to most diffusion-model variants. We define influence through a local diffusion score discrepancy and derive an efficiently computable gradient similarity score to estimate it. Next, we show that this score can be distilled into the diffusion model's own forward representations, yielding \studentfull{} (\student{}, Sec.~\ref{sec:method-student}), further removing the remaining backward passes during attribution and compacting the embeddings into a low-dimensionality space. Additional overview and a block diagram can be found in Appendix~\ref{app:overview}.

\paragraph{Notation.}
Consider a general diffusion model under common principles~\citep{lai2025principles}, parametrized by $\theta$, with a learnt score $s_\theta$ obtained from a neural network $F_\theta$. The latter is pre-trained with a reconstruction loss $L$ that involves noise scales $\sigma$, noise vectors $\mathbf{n}$, and training instances $\mathbf{z}$. Each instance consists of a data item $\mathbf{x}$ and, optionally, a conditioning signal $\mathbf{c}$, i.e.,~$\mathbf{z}=\{\mathbf{x},\mathbf{c}\}$. We denote the training set by $\mathcal{Z}=\{\mathbf{z}_1,\dots,\mathbf{z}_{|\mathcal{Z}|}\}$. At inference time, the model generates an item $\hat{\mathbf{x}}$ by iterative denoising along a decreasing schedule of noise scales $\sigma_{\max}=\sigma_T>\dots>\sigma_1=\sigma_{\min}$ using conditioning $\hat{\mathbf{c}}$. We write $\hat{\mathbf{z}}=\{\hat{\mathbf{x}},\hat{\mathbf{c}}\}$ for the resulting attribution query. We also use $\mathbf{x}_{\sigma,\mathbf{n}}$ and $\hat{\mathbf{x}}_{\sigma,\mathbf{n}}$ to denote the forward noising of a training or a generated item, respectively (see Appendix~\ref{app:theoret-bckgrd}).

\subsection{Local Score Discrepancy}
\label{sec:method-teacher}

TDA asks how removing a training sample $\mathbf{z}_i$ changes a model's behavior at a given query $\hat{\mathbf{z}}$. For a generative model, this effect is naturally characterized by how the generation produced by the leave-one-out model $F_{\theta^{\setminus i}}$ differs from the original generation, where $\theta^{\setminus i}$ denotes the parameters obtained by retraining from scratch without $\mathbf{z}_i$. To measure this effect, existing work usually relies on indirect counterfactual proxies, such as changes in the training loss~\citep{georgiev2023journey,zheng2024intriguing,mlodozeniec2025influence}. In a diffusion model, however, generation is driven by the learned score field~\citep{song2020score,song2020improved}. Perturbations to the training distribution induce changes in the optimal score field, which propagate through the sampling dynamics to affect generated samples~\citep{scarvelis2025sensitivityanalysisdiffusionmodels}. We therefore quantify the removal effect in this space, comparing the score functions of $s_\theta$ and $s_{\theta^{\setminus i}}$ along noisy states originating from the fixed query/generated item. To do so, we consider a Monte Carlo estimate using a sample of noise scales $\sigma$ and noise realizations $\mathbf{n}$, and study the score discrepancy at a given anchor point $\hat{\mathbf{z}}$ (the query/generated item). We define the \textit{score discrepancy at $\hat{\mathbf{z}}$} caused by $\mathbf{z}_i$ as
\begin{equation*}
\tau_\theta(\mathbf{z}_i;\hat{\mathbf{z}})
:=\mathbb{E}_{\mathbf{\sigma},\mathbf{n}}\!\left[
\left\lVert
s_\theta\left(\hat{\mathbf{x}}_{\sigma,\mathbf{n}};\sigma,\hat{\mathbf{c}}\right)
- s_{\theta^{\setminus i}}\left(\hat{\mathbf{x}}_{\sigma,\mathbf{n}};\sigma,\hat{\mathbf{c}}\right)
\right\rVert_2^2
\right],
\end{equation*}
where $s_\theta$ and $s_{\theta^{\setminus i}}$ denote the scores learned by the models using $\mathcal{Z}$ and $\mathcal{Z}\setminus\mathbf{z}_i$, respectively. Intuitively, a larger $\tau_\theta(\mathbf{z}_i;\hat{\mathbf{z}})$ indicates that removing $\mathbf{z}_i$ shifts the learned score field more strongly around the query $\hat{\mathbf{z}}$.

Because the score estimate is common to most diffusion variants, this definition is agnostic to any particular parameterization: it applies whenever the generative model provides a noise- or a time-indexed score, including DDPM~\citep{ho2020denoising}, EDM~\citep{karras2022elucidating}, and flow-matching models~\citep{lipman2022flow,liu2022flow,albergo2022stochastic}. Moreover, for the common parameterizations, the score estimate is an affine function of the raw network output with schedule-dependent coefficients~\citep{lai2025principles}, so the discrepancy can be evaluated on model outputs directly:
\begin{equation}
\label{eq:tau-output}
\tau_\theta(\mathbf{z}_i;\hat{\mathbf{z}})
=\mathbb{E}_{\sigma,\mathbf{n}}\!\left[
w_\sigma\left\lVert
F_\theta\big(\hat{\mathbf{x}}_{\sigma,\mathbf{n}};\sigma,\hat{\mathbf{c}}\big)
-F_{\theta^{\setminus i}}\big(\hat{\mathbf{x}}_{\sigma,\mathbf{n}};\sigma,\hat{\mathbf{c}}\big)
\right\rVert_2^2
\right],
\end{equation}
where the weight $w_\sigma$ depends only on the noise scale $\sigma$ and fully encodes the diffusion variant. Appendix~\ref{app:method-weights} derives Eq.~\ref{eq:tau-output} and instantiates $w_\sigma$ for DDPM, EDM, and flow matching.

\paragraph{Linearization: from output differences to gradient similarity.}
Directly evaluating Eq.~\ref{eq:tau-output} requires retraining a new leave-one-out model $F_{\theta^{\setminus i}}$ for every training sample. Since this is computationally infeasible, we next introduce a practical instantiation tailored at more realistic deployments. We start by considering a first-order Taylor expansion around the trained parameters $\theta$:
\begin{equation*}
F_{\theta^{\setminus i}}\big(\hat{\mathbf{x}}_{\sigma,\mathbf{n}};\sigma,\hat{\mathbf{c}}\big)
\approx F_\theta\big(\hat{\mathbf{x}}_{\sigma,\mathbf{n}};\sigma,\hat{\mathbf{c}}\big)
+\nabla_\theta F_\theta\big(\hat{\mathbf{x}}_{\sigma,\mathbf{n}};\sigma,\hat{\mathbf{c}}\big)\Delta\theta_i,
\qquad \Delta\theta_i:=\theta^{\setminus i}-\theta,
\end{equation*}
where $\nabla_\theta F_\theta\big(\hat{\mathbf{x}}_{\sigma,\mathbf{n}};\sigma,\hat{\mathbf{c}}\big)\in\mathbb{R}^{D \times P}$ is the Jacobian of the model output with respect to the parameters, $D$ is the output dimension, and $\Delta\theta_i\in\mathbb{R}^{P}$ is the parameter change induced by removing $\mathbf{z}_i$. Substituting into Eq.~\ref{eq:tau-output} gives
\begin{equation}
\label{eq:tau-linearized}
\tau_\theta(\mathbf{z}_i;\hat{\mathbf{z}})
\approx \mathbb{E}_{\sigma,\mathbf{n}}\!\left[
w_\sigma
\left\lVert\nabla_\theta F_\theta\big(\hat{\mathbf{x}}_{\sigma,\mathbf{n}};\sigma,\hat{\mathbf{c}}\big)\Delta\theta_i\right\rVert_2^2\right].
\end{equation}

Since the model output is high-dimensional, we follow TRAK~\citep{park2023trak} and DAS~\citep{lin2025diffusion} to approximate its change through a scalar regression model whose output is the sum of the model output coordinates, and then estimate the counterfactual parameter change $\Delta\theta_i$ by a one-step Newton update on the removed sample. We consider the scalar regression model producing the summed output $\langle\mathbf{1},F_\theta\rangle$ and, with such output, define the model's parameter gradient as
\begin{equation}
\label{eq:probe-grad}
g(\mathbf{z};\sigma,\mathbf{n})
:=\nabla_\theta\big\langle \mathbf{1},
F_\theta(\mathbf{x}_{\sigma,\mathbf{n}};\sigma,\mathbf{c})\big\rangle\;\in\;\mathbb{R}^{P}.
\end{equation}
For simplicity, we will omit $(\sigma,\mathbf{n})$ from $g$ and related quantities when there is no ambiguity. Next, to estimate the parameter change $\Delta\theta_i$, we apply the Newton's method~\citep{pregibon1981logistic} to the scalar regression model, where a one-step Newton update approximation gives the closed form
\begin{equation}
\label{eq:newton}
\Delta\theta_i\approx-\frac{r_i}{1-h_i}\,\mathbf{G}^{-1}g(\mathbf{z}_i),
\qquad
\mathbf{G}=\mathbb{E}_{\mathbf{z},\sigma,\mathbf{n}}\big[g(\mathbf{z};\sigma,\mathbf{n})\,g(\mathbf{z};\sigma,\mathbf{n})^{\top}\big],
\end{equation}
where $\mathbf{G}$ is the curvature of the regression model, the leverage $h_i=g(\mathbf{z}_i)^{\top}\mathbf{G}^{-1}g(\mathbf{z}_i)\in[0,1)$, and the training residual $r_i=\langle\mathbf{1},\,F_\theta(\mathbf{x}_{\sigma,\mathbf{n}};\sigma,\mathbf{c})-\mathbf{y}\rangle$, evaluated at $\mathbf{z}_i$, with $\mathbf{y}$ representing the diffusion variant's regression target \citep[for example, $\mathbf{y}=\mathbf{n}$ for DDPM and $\mathbf{y}=\mathbf{x}$ for EDM, cf.][]{park2023trak,lin2025diffusion}. For ease of calculation, we drop the factor $r_i/(1-h_i)$, as our ablations confirm that such terms are negligible to the ranking (Sec.~\ref{sec:res-ablation}). Dropping such factor and substituting Eqs.~\ref{eq:probe-grad} and \ref{eq:newton} into Eq.~\ref{eq:tau-linearized} yields
\begin{equation}
\label{eq:tau-probe}
\tau_\theta(\mathbf{z}_i,\hat{\mathbf{z}})
\approx
\mathbb{E}_{\sigma,\mathbf{n}}\Big[\,
w_\sigma\,
\big\langle g(\hat{\mathbf{z}};\sigma,\mathbf{n}), \mathbf{G}^{-1}
g(\mathbf{z}_i;\sigma,\mathbf{n}) \big\rangle^{2}\Big].
\end{equation}

\paragraph{Aligned draws and inference.}
In practice, the expectation in Eq.~\ref{eq:tau-probe} is estimated by Monte Carlo sampling, with $\sigma$ drawn from the diffusion sampler's own schedule. Although the Newton update imposes no restriction on the choice of noise scale $\sigma$ or noise realization $\mathbf{n}$, \citet{serra2026training} observed that aligning diffusion scheduler timesteps can dramatically improve attribution performance. Hence, to make the scores $\tau_\theta(\mathbf{z}_i;\hat{\mathbf{z}})$ directly comparable across training samples, we follow such strategy and use the same set of noise scales $\sigma$ and noise realizations $\mathbf{n}$ for all the queries, with each draw $(\sigma,\mathbf{n})$ being shared among queries on both sides of the kernel in Eq.~\ref{eq:tau-probe}. Specifically, we use $M$ noise scales of the sampler's Karras grid, each paired with a single noise vector drawn once under a fixed seed and shared across all samples and queries. (we study how $M$ trades off cost and accuracy in Sec.~\ref{sec:res-efficiency}).

\paragraph{Layer-wise kernel estimation with Kronecker factorization.}
Evaluating Eq.~\ref{eq:tau-probe} requires computing the inner product kernels $\langle g(\hat{\mathbf{z}}), \mathbf{G}^{-1} g(\mathbf{z}_i)\rangle$ for all pairs. Most existing works compress those vectors with random projections, introducing projection-induced distortion on kernel estimation~\citep{park2023trak,zheng2024intriguing,lin2025diffusion}. Following K-FAC Influence~\citep{mlodozeniec2025influence}, we instead approximate $\mathbf{G}$ blockwise, per layer, with Kronecker-factored approximate curvature~\citep[K-FAC;][]{martens2015optimizing,george2018fast}. For a linear layer $l$ with input activations $\mathbf{a}_{l}$ and pre-activation output gradients $\mathbf{b}_{l}$, the layer gradient can be decomposed as $g_l(\mathbf{z})=\mathbf{b}_{l}\,\mathbf{a}_{l}^{\top}$, which gives the following Kronecker factorization:
\begin{equation}
\label{eq:kfac}
\mathbf{G}_l \approx \mathbf{A}_l \otimes \mathbf{B}_l
\quad\Longrightarrow\quad
\big\langle g_l(\hat{\mathbf{z}}),\mathbf{G}_l^{-1} g_l(\mathbf{z}_i)\big\rangle
\approx \big(\hat{\mathbf{a}}_{l}^{\top} \mathbf{A}_l^{-1} \mathbf{a}_{l,i}\big)\big(\hat{\mathbf{b}}_{l}^{\top} \mathbf{B}_l^{-1} \mathbf{b}_{l,i}\big),
\end{equation}
where $\otimes$ is the Kronecker product, $\mathbf{A}_l=\mathbb{E}_{\mathbf{z},\sigma,\mathbf{n}}[\mathbf{a}_{l} \mathbf{a}_{l}^{\top}]$, $\mathbf{B}_l=\mathbb{E}_{\mathbf{z},\sigma,\mathbf{n}}[\mathbf{b}_{l} \mathbf{b}_{l}^{\top}]$, and $(\hat{\mathbf{a}}_{l},\hat{\mathbf{b}}_{l})$ and $(\mathbf{a}_{l,i},\mathbf{b}_{l,i})$ denote the factors of the query and of $\mathbf{z}_i$, respectively. A similar factorization applies to other common layer types in diffusion models, including convolutional or transposed convolutional layers (we provide the corresponding derivations in Appendix~\ref{app:layer-extensions}).

\paragraph{Normalization.}
In our implementation, per-sample and per-layer gradient factors are L2-normalized before accumulation, which is a common practice in gradient-similarity attribution that prevents samples with large gradient norms from dominating the scores~\citep{barshan2020relatif,hanawa2020evaluation,xia2024less}. Without normalization, gradients evaluated at larger $\sigma$, where the inputs contain more noise, can dominate the aggregation across noise levels and make the resulting rankings sensitive to the sampled noise scales. For a similar reason, we drop the weight $w_\sigma$: after normalization, each sampled noise scale contributes a standardized influence value with a common scale, and reintroducing a schedule-dependent multiplier would disturb this standardization. Putting everything together, our final attribution score becomes
\begin{equation}
\label{eq:tau-kfac}
\tau_\theta(\mathbf{z}_i;\hat{\mathbf{z}})
\approx \mathbb{E}_{\sigma,\mathbf{n}}\!\left[
\left(\sum_l
\big(\hat{\underline{\mathbf{a}}}_l^\top\underline{\mathbf{A}}_l^{-1}\underline{\mathbf{a}}_{l,i}\big)
\big(\hat{\underline{\mathbf{b}}}_l^\top\underline{\mathbf{B}}_l^{-1}\underline{\mathbf{b}}_{l,i}\big)
\right)^2\right],
\end{equation}
where $\underline{\mathbf{A}}_l=\mathbb{E}_{\mathbf{z},\sigma,\mathbf{n}}[\underline{\mathbf{a}}_{l} \underline{\mathbf{a}}_{l}^{\top}]$, $\underline{\mathbf{B}}_l=\mathbb{E}_{\mathbf{z},\sigma,\mathbf{n}}[\underline{\mathbf{b}}_{l} \underline{\mathbf{b}}_{l}^{\top}]$, and the underline indicates L2 normalization.

\paragraph{Relation to existing gradient-based methods.}
\teacher{} belongs to the broad family of gradient-based attribution methods, which includes TRAK~\citep{park2023trak}, D-TRAK~\citep{zheng2024intriguing}, DAS~\citep{lin2025diffusion}, and K-FAC Influence~\citep{mlodozeniec2025influence}. \teacher differs mainly in the \emph{attribution target} being differentiated and the \emph{curvature model} used to precondition its gradient. Furthermore, it offers a more general attribution framework. Note that, for a DDPM noise predictor, Eq.~\ref{eq:tau-output} reduces to the expected squared difference of noise predictions that motivates DAS. However, DAS obtains it from a KL divergence specialized to the DDPM reverse process and the loss $\mathcal{L}_\text{simple}$, whereas \teacher obtains it from the score discrepancy, which does not rely on the probabilistic parameterization of DDPM nor the specific loss being used, and hence extends to EDM and flow matching. Beyond this formulation difference, the methods differ in five design choices: (i)~\emph{attribution target}: \teacher differentiates the summed model output, whereas DAS and K-FAC Influence differentiate the denoising loss and D-TRAK differentiates the squared output norm; (ii)~\emph{corruption sampling}: \teacher evaluates query and candidates at the same $(\sigma,\mathbf{n})$, with $\sigma$ drawn from the inference schedule, whereas D-TRAK, DAS and K-FAC Influence corrupt the samples independently; (iii)~\emph{gradient normalization}: \teacher normalizes the two Kronecker factors of each layer before aggregating the layerwise kernels, whereas DAS normalizes the whole projected gradient at each timestep and D-TRAK and K-FAC Influence use unnormalized gradients; (iv)~\emph{Newton corrections}: \teacher omits the residual and leverage terms $r_i$ and $1-h_i$ of Eq.~\ref{eq:newton}, whereas D-TRAK retains a residual-type factor $\mathcal{Q}$ and DAS retains both; and (v)~\emph{curvature estimation}: \teacher forms empirical-Fisher K-FAC factors from observed activations and output gradients, whereas D-TRAK and DAS use a ridge-damped Gauss--Newton kernel of randomly projected gradients, which introduces projection-induced distortion, while K-FAC Influence estimates a Monte Carlo generalized Gauss--Newton matrix by resampling the regression target. Sec.~\ref{sec:res-ablation} ablates i--v and Appendix~\ref{sec:related-work} gives a broader literature review, including non-gradient-based attribution methods.

\subsection{Distilling Attribution into Forward Embeddings}
\label{sec:method-student}

While \teacher{} avoids per-sample gradient storage, each attribution score still requires a forward--backward pass per compared sample and implies a potentially high dimensionality (number of model parameters). To further reduce such costs, we additionally propose \studentfull{} (\student{}), which eliminates the need for backward passes entirely and projects to a lower dimensionality $d$. In this setting, \student{} becomes a student method distilled from its teacher, \teacher{} (hereafter referred to as the \emph{teacher}). It learns an embedding $\mathbf{e}\in\mathbb{R}^{d}$ from the diffusion model's internal activations, obtained from a single forward pass, such that the embedding similarity reproduces the \teacher{}'s attribution ranking at dimensions as low as $d=768$; orders of magnitude below the gradient representations they replace. In principle, any attribution method could serve as the teacher, but \teacher{}'s factorized, per-layer scores make distillation particularly efficient.

\paragraph{Embedding model architecture.}
\student uses no external encoder and its input consists of internal representations already computed by the diffusion model $F_\theta$. Let $\mathbf{h}_{k}(\mathbf{z};\sigma,\mathbf{n})$ denote the token-wise residual-stream activations after the $k$-th of the $K$ transformer blocks of $F_\theta$ when processing the noised input $\mathbf{x}_{\sigma,\mathbf{n}}$. An embedding $\mathbf{e}_i$ is computed from the blockwise activations sum through a learned attention pooling over tokens, followed by a small MLP (all in a single forward pass):
\begin{equation}
\label{eq:embeddings}
\mathbf{e}_i = E_\phi(\mathbf{z}_i;\sigma,\mathbf{n})
=\mathrm{MLP}\left(\mathrm{AttnPool}\left(\sum_{k=1}^{K}\mathbf{h}_k(\mathbf{z}_i;\sigma,\mathbf{n})\right)\right).
\end{equation}
The score between two embeddings $\mathbf{e}_i$ and $\mathbf{e}_j$ is their cosine similarity $\rho_{ij} = \rho_\phi(\mathbf{z}_i,\mathbf{z}_j;\sigma,\mathbf{n}) := \mathbf{e}_i^\top\mathbf{e}_j/(\lVert\mathbf{e}_i\rVert\,\lVert\mathbf{e}_j\rVert)$.

\paragraph{Batched pair-wise score distillation.}
The prior distilled attribution of \citet{wang2026fast} learns its student from an unlearning-based teacher. Hence, annotating the influence of a training sample requires a dedicated unlearning run per training sample, making it substantially more expensive to expand supervision coverage on the training side than to add query samples. The supervision set is therefore curated offline for a fixed collection of pairs, and limited to the coverage budgeted before training. \teacher{} removes this asymmetry, as it requires only one single forward--backward pass per training or query sample. For each training data batch $\mathcal{B}=\{\mathbf{z}_1,\dots,\mathbf{z}_B\}$, we draw a single shared $(\sigma,\mathbf{n})$ under the aligned regime of Sec.~\ref{sec:method-teacher} and compute, in one forward--backward pass, both the within-batch pairwise \student{} scores $\rho_{ij}$ and the teacher's within-batch pairwise scores $\tau_{ij}:=\tau_\theta(\mathbf{z}_i;\mathbf{z}_j)$. Compared to offline annotation, this has two advantages: (i)~\emph{coverage}: every step supervises a set of $B^2$ pairs, so the supervised pairs accumulate throughout training rather than being fixed in advance; and (ii)~\emph{no annotation overhead}: supervision is computed when it is consumed and is never materialized or stored. We provide the distillation algorithm summary in Appendix~\ref{app:overview}.

\paragraph{Ranking objective.}
Attribution is usually consumed as per-query rankings~\citep{ghorbani2019data,xia2024less,wang2026fast}, so we distill rankings rather than raw score values. Pointwise regression of scores or normalized ranks, as in \citet{wang2026fast}, weights every candidate equally and under-constrains the top of the ranking. In contrast, we adopt a top-heavy pairwise objective from the LambdaLoss family~\citep{burges2005learning,wang2018lambdaloss}. Within a batch, we treat each sample $i$ as an anchor and every other pair $(j,k)$ as candidates. Let $r_{ij}\in\{1,\dots,B{-}1\}$ denote the rank of candidate $j$ under the teacher's row $\tau_{i:}$ in descending order, and let $D(r)=1/\log_2(1{+}r)$ denote the DCG discount~\citep{jarvelin2002cumulated}. \student{} is trained with the weighted pairwise binary cross-entropy
\begin{equation}
\label{eq:lambdarank}
\mathcal{L}_{\mathrm{rank}}(\rho,\tau)
=\frac{1}{Z}\sum_{i}\sum_{\substack{j<k\\ j,k\neq i}}
\big|D(r_{ij})-D(r_{ik})\big| \cdot
\mathrm{BCE}\Big(\text{sigmoid}\big(\alpha\cdot(\rho_{ij}-\rho_{ik})\big),
\mathds{1}\big[\tau_{ij}>\tau_{ik}\big]\Big),
\end{equation}
where $\alpha>0$ is a learnable global logit scale and $Z=\sum |D(r_{ij})-D(r_{ik})|$ normalizes the weights. The DCG discount concentrates the loss on pairs near the top of the teacher's ranking, matching the top-k fidelity by which attribution is evaluated. Teacher quantities ($\tau$ and $r$) are detached, so only the embedding model receives gradients. Because Eq.~\ref{eq:lambdarank} depends on embedding score differences, it is invariant to any per-anchor shift of $\rho$, leaving the logit scale $\alpha$ as the sole calibration parameter. Following the logit-scale treatment of CLIP~\citep{radford2021learning}, we parameterize it on a log scale.

\paragraph{Aligned multi-draw inference.}
Similar to when computing the expectation with \teacher{} (Eq.~\ref{eq:tau-probe}), we here also use multiple aligned draws at inference time and average $\rho$ scores over $M$ draws $\{(\sigma_m,\mathbf{n}_m)\}_{m=1}^{M}$ shared between the query and every training sample. Notice that the inference procedure can be efficient, as training-side embeddings can be computed once and cached. Each query requires only $M$ forward passes at inference time (compared to the $|\mathcal{Z}|\cdot M$ passes required for the train set). Because the query embedding is constructed from the model's own activations, attribution is also available during generation itself ($M$ is in the order of the number of diffusion steps and $E_\phi$ is much smaller than $F_\theta$; we ablate $M$ in Sec.~\ref{sec:res-efficiency} and provide architecture details in Appendix~\ref{app:exp-details}).

%% file: sections/experiments.tex
\section{Experiments and Results}
\label{sec:experiments}

\begin{figure}[t]
\centering
\includegraphics[width=\linewidth]{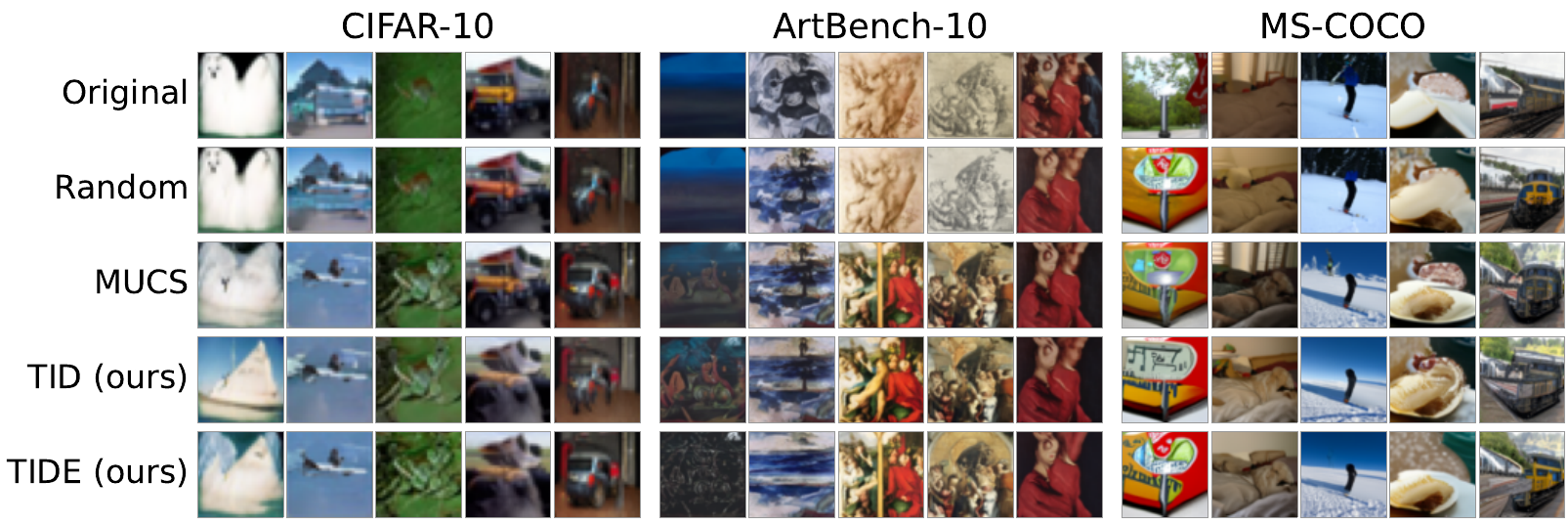}
\vspace{-15pt}
\caption{Qualitative comparison of five target queries from each dataset. Top row: original generations from the full-data model. Rows below: regenerations from the same initial noise after removing the samples attributed by each method. Random removal leaves regenerations nearly unchanged, whereas removal by MUCS, \teacher{}, and \student{} induces a visible drift.}
\label{fig:qualitative}
\end{figure}

\subsection{Methodology}

Following MUCS~\citep{serra2026training}, we evaluate attribution quality using \emph{seed consistency}: diffusion models retrained after random data removal generate nearly identical images from the same initial noise, whereas removing samples that support a generation causes it to change. For each dataset and seed, we (i)~train a model on the full dataset and generate a fixed set of queries; (ii)~score all training samples for each query and remove the union of the per-query top-$2\%$; (iii)~retrain the model from scratch on the remaining data; and (iv)~regenerate the queries using the same initial noise. Better attribution methods identify more influential samples and therefore induce greater deviation from the original generations.

We measure similarity between each original and regenerated image using SSIM~\citep{wang2004image}, cosine similarity in the SSCD copy-detection space~\citep{pizzi2022self}, LPIPS~\citep{zhang2018unreasonable}, and cosine similarity in the CLIP semantic space~\citep{radford2021learning}. For each metric, we compare the similarities induced by a method's removals with those from an equally sized random-removal control, and summarize their separation using the area under the ROC curve (AUC). An AUC of 0.5 indicates performance equivalent to random removal, whereas 1 indicates perfect separation. We report the AUC for each metric, their mean $\mu$, and the mean $\pm$ standard error over six different runs.

\paragraph{Data and models.}
We evaluate across diverse settings, including unconditional CIFAR-10~\citep{alex2009learning}, class-conditional ArtBench-10~\citep{liao2022artbench}, and also text-conditioned MS-COCO~\citep{lin2014microsoft} using CLIP embeddings. For each dataset, we train a DiT backbone~\citep{peebles_scalable_2023} under the EDM formulation~\citep{karras2022elucidating}. Full training, sampling, and evaluation details, along with an analysis of removal budgets, are provided in Appendix~\ref{app:exp-details}.

\paragraph{Baselines.}
We compare against representative TDA methods from three broad families. \emph{Model-agnostic} baselines retrieve samples by cosine similarity between CLIP~\citep{radford2021learning} or DINO~\citep{oquab2023dinov2} image embeddings. \emph{Gradient-based} baselines include D-TRAK~\citep{zheng2024intriguing} and DAS~\citep{lin2025diffusion}, which use (preconditioned) gradient similarity in a randomly projected parameter space. Although these methods were originally developed for DDPMs only, we adapt them to the EDM formulation while preserving their underlying design principles. As for \emph{unlearning-based} baseline we consider MUCS~\citep{serra2026training}, which fine-tunes the model to unlearn each query and measures the resulting loss changes. We run all methods under the same evaluation protocol, using baseline hyperparameters adapted from their original settings (Appendix~\ref{app:exp-details}).

\begin{table}[t]
\centering
\caption{Counterfactual AUC per similarity metric and their average $\mu$, reported as the mean over six seeds $\pm$ the standard error. Categories: MA = model-agnostic, G = gradient-based, U = unlearning-based, D = distillation-based. Best per column in \textbf{bold}, second best \underline{underlined}.}
\label{tab:main}
\vspace{1.5mm}
\newcommand{\se}[1]{{\fontsize{6.5}{7.5}\selectfont ~$\pm$~#1}}
\setlength{\tabcolsep}{8pt}
\resizebox{1\columnwidth}{!}{%
\begin{tabular}{clcccccc}
\toprule
& Approach & Categ. & SSIM & SSCD & LPIPS & CLIP & $\mu$ \\
\midrule
\multirow{7}{*}{\rotatebox[origin=c]{90}{CIFAR-10}}
& CLIP & MA & 0.598\se{0.011} & 0.585\se{0.008} & 0.610\se{0.013} & 0.661\se{0.010} & 0.613\se{0.007} \\
& DINO & MA & 0.604\se{0.018} & 0.616\se{0.017} & 0.638\se{0.021} & 0.665\se{0.027} & 0.631\se{0.019} \\
& D-TRAK & G & 0.691\se{0.037} & 0.649\se{0.033} & 0.703\se{0.026} & 0.640\se{0.042} & 0.671\se{0.033} \\
& DAS & G & 0.760\se{0.031} & 0.753\se{0.036} & 0.789\se{0.040} & 0.746\se{0.035} & 0.762\se{0.035} \\
& MUCS & U & \textbf{0.864}\se{0.025} & \textbf{0.854}\se{0.023} & \textbf{0.908}\se{0.015} & \textbf{0.857}\se{0.028} & \textbf{0.870}\se{0.021} \\
& \teacher{} (ours) & G & 0.836\se{0.031} & 0.815\se{0.031} & 0.872\se{0.023} & 0.818\se{0.027} & 0.835\se{0.027} \\
& \student{} (ours) & D & \underline{0.856}\se{0.018} & \underline{0.844}\se{0.017} & \underline{0.901}\se{0.013} & \underline{0.832}\se{0.013} & \underline{0.858}\se{0.013} \\
\midrule
\multirow{7}{*}{\rotatebox[origin=c]{90}{ArtBench-10}}
& CLIP & MA & 0.575\se{0.036} & 0.575\se{0.031} & 0.588\se{0.036} & 0.621\se{0.018} & 0.590\se{0.029} \\
& DINO & MA & 0.558\se{0.038} & 0.553\se{0.028} & 0.573\se{0.032} & 0.586\se{0.034} & 0.567\se{0.032} \\
& D-TRAK & G & 0.752\se{0.035} & 0.707\se{0.030} & 0.745\se{0.031} & 0.677\se{0.032} & 0.720\se{0.029} \\
& DAS & G & 0.816\se{0.020} & 0.794\se{0.025} & 0.819\se{0.021} & 0.787\se{0.031} & 0.804\se{0.022} \\
& MUCS & U & \underline{0.867}\se{0.013} & 0.832\se{0.018} & 0.869\se{0.015} & \underline{0.831}\se{0.012} & \underline{0.850}\se{0.012} \\
& \teacher{} (ours) & G & \textbf{0.905}\se{0.007} & \textbf{0.870}\se{0.011} & \textbf{0.918}\se{0.006} & \textbf{0.861}\se{0.019} & \textbf{0.889}\se{0.009} \\
& \student{} (ours) & D & \underline{0.867}\se{0.009} & \underline{0.835}\se{0.012} & \underline{0.872}\se{0.018} & 0.825\se{0.020} & \underline{0.850}\se{0.011} \\
\midrule
\multirow{7}{*}{\rotatebox[origin=c]{90}{MS-COCO}}
& CLIP & MA & 0.648\se{0.022} & 0.678\se{0.027} & 0.691\se{0.028} & \underline{0.799}\se{0.036} & 0.704\se{0.026} \\
& DINO & MA & 0.683\se{0.019} & 0.712\se{0.032} & 0.718\se{0.036} & \textbf{0.811}\se{0.027} & 0.731\se{0.027} \\
& D-TRAK & G & 0.696\se{0.023} & 0.704\se{0.033} & 0.746\se{0.035} & 0.698\se{0.031} & 0.711\se{0.028} \\
& DAS & G & 0.718\se{0.016} & 0.712\se{0.033} & 0.761\se{0.026} & 0.739\se{0.035} & 0.733\se{0.024} \\
& MUCS & U & 0.722\se{0.019} & 0.717\se{0.021} & 0.757\se{0.025} & 0.755\se{0.039} & 0.738\se{0.022} \\
& \teacher{} (ours) & G & \textbf{0.773}\se{0.019} & \textbf{0.757}\se{0.022} & \textbf{0.813}\se{0.029} & 0.788\se{0.029} & \textbf{0.783}\se{0.020} \\
& \student{} (ours) & D & \underline{0.738}\se{0.025} & \underline{0.727}\se{0.028} & \underline{0.781}\se{0.025} & 0.748\se{0.012} & \underline{0.748}\se{0.018} \\
\bottomrule
\end{tabular} }
\vspace{-10pt}
\end{table}

\subsection{Comparison with Existing Methods}
\label{sec:res-existing}

Figure~\ref{fig:qualitative} illustrates the levels of degradation induced by removing samples identified by different methods. Under random removal, the regenerations are nearly indistinguishable from the originals, which illustrates the seed consistency that the evaluation relies on. Under \teacher{} and \student{}, the regenerations visibly drift in composition, style, or content, indicating that the removed samples indeed supported the queries. Additional qualitative results and discussion can be found in Appendix~\ref{app:qual-topbottom}.

Table~\ref{tab:main} reports the counterfactual AUCs for the considered methods. \teacher{} attains the overall highest accuracy among the attribution methods: it surpasses MUCS on ArtBench-10 and MS-COCO while performing almost on par with it on CIFAR-10. The other existing gradient-based methods perform worse on all three datasets. \student{}, in turn, closely approaches the accuracy of its teacher \teacher{}, while requiring only forward passes and a lower dimensionality (we quantify this efficiency advantage in the following section). Additional quantitative results are available in Appendices~\ref{app:sensitivity} and~\ref{app:ratio}.

\subsection{Efficiency Trade-offs between TDA Methods}
\label{sec:res-efficiency}

\begin{figure}[t]
\centering
\includegraphics[width=0.98\linewidth]{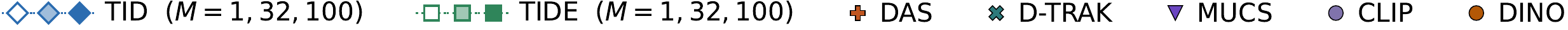}\\[3pt]

\begin{minipage}[t]{0.362\linewidth}
\includegraphics[width=\linewidth]{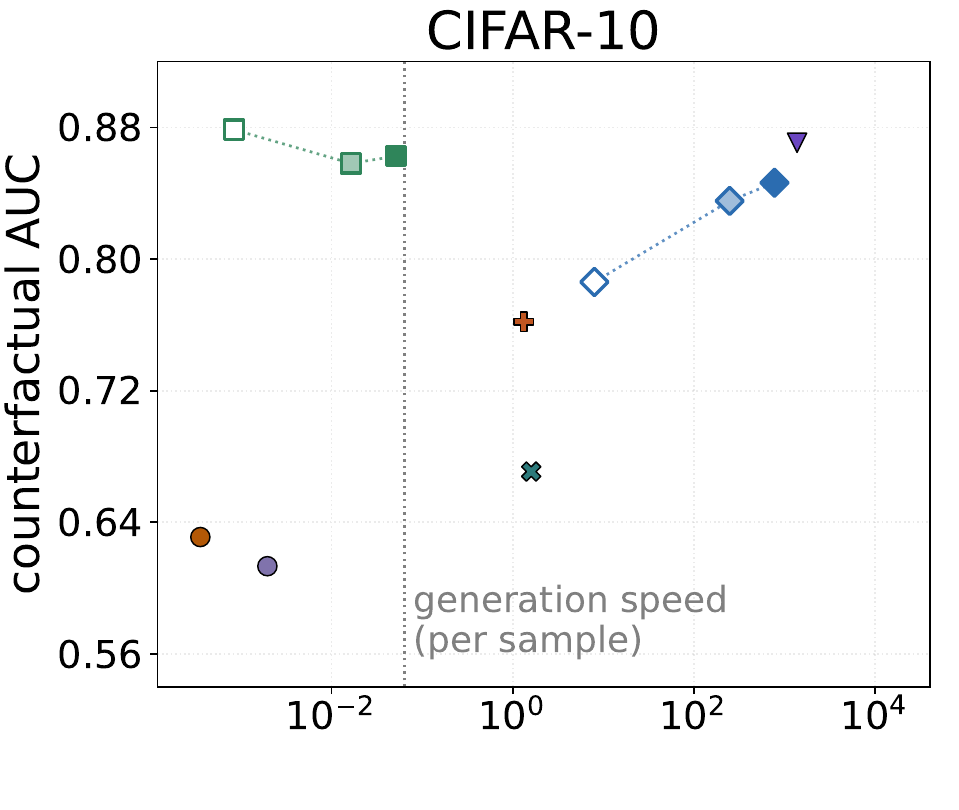}
\end{minipage}\hfill
\begin{minipage}[t]{0.317\linewidth}
\includegraphics[width=\linewidth]{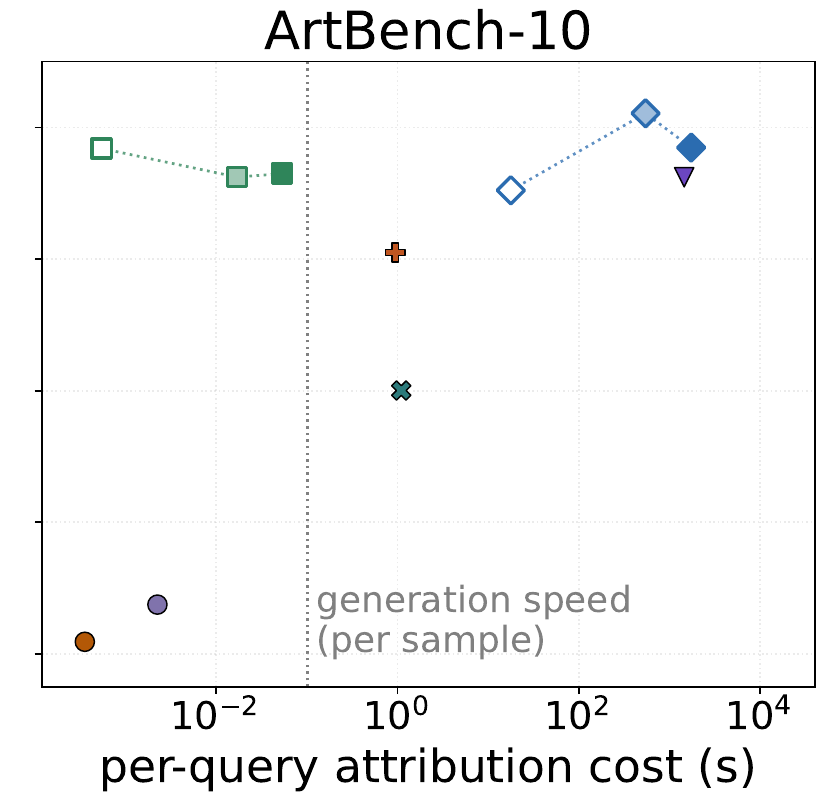}
\end{minipage}\hfill
\begin{minipage}[t]{0.317\linewidth}
\includegraphics[width=\linewidth]{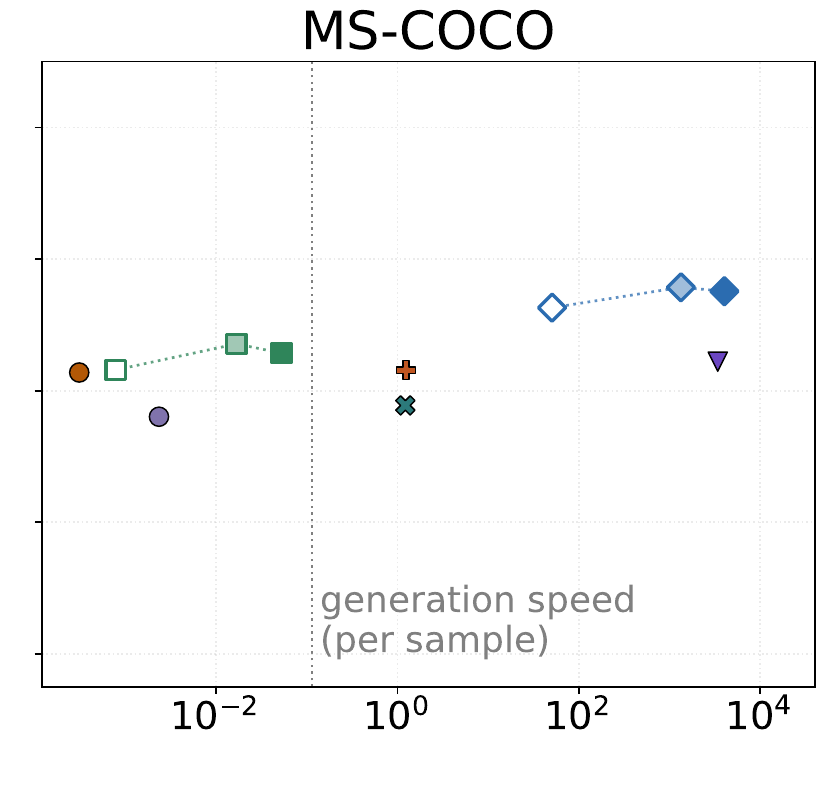}
\end{minipage}
\vspace{-10pt}
\caption{Attribution accuracy versus warm-cache per-query cost on CIFAR-10 (left), ArtBench (middle), and MS-COCO (right). The horizontal axis is logarithmic; cacheable training-side artifacts are precomputed. Markers with different fills connected by dotted lines trace the variation with $M$. }
\label{fig:efficiency}
\end{figure}

Attribution costs vary significantly (even at similar levels of performance, the methods in Table~\ref{tab:main} differ by orders of magnitude in their per-query attribution cost). To characterize this efficacy--efficiency trade-off, we compare all methods on a common accuracy--cost plane. Figure~\ref{fig:efficiency} plots counterfactual AUCs against per-query cost. The methods essentially fall into two regimes. CLIP, DINO, DAS, and \student{} process only the query at attribution time and complete each attribution in microseconds to fractions of a second. In contrast, \teacher{} and MUCS repeat training-side computation for every query. \student{} alone occupies the upper-left region of the plane, combining the cost and storage profile of embedding-based methods with the accuracy of the expensive gradient- or unlearning-based ones. We note that \student{} can attribute a sample faster than the model generates it (vertical dotted lines in Figure~\ref{fig:efficiency}). Appendix~\ref{app:efficiency-details} further provides an asymptotic cost analysis of each method, together with the one-off training-side and on-disk costs.

\subsection{Dissecting the Design of \teacher{}}
\label{sec:res-ablation}

\teacher{} differs from existing gradient-based estimators along several dimensions (Sec.~\ref{sec:method-teacher}). We here isolate these design choices through five research questions. Specifically, we vary the scalar target being differentiated (RQ1), the corruption draws used to evaluate the query and training samples (RQ2), the normalization of per-sample gradients before kernel construction (RQ3), the terms retained from the exact counterfactual-retraining expansion (RQ4), and the curvature model used for preconditioning (RQ5). Table~\ref{tab:teacher-ablation} reports the corresponding ablations on CIFAR-10.

\paragraph{RQ1: How much does the choice of attribution target matter?}
\teacher{} differs from existing gradient-based methods in its attribution target and curvature model. We first isolate the target by fixing the curvature model to that of \teacher{} and comparing its summed model output against the denoising loss $\omega(\sigma)\lVert\mathbf{x}-F_\theta\rVert_2^2$ (e.g.,~TRAK) and squared output norm $\lVert F_\theta\rVert_2^2$ (e.g.,~D-TRAK). We observe that both alternatives substantially reduce attribution accuracy, with the denoising loss causing the largest drop.

\paragraph{RQ2: Should corruption draws be aligned?}
Each gradient depends on the corruption draw at which it is evaluated. Although a similarity between gradients computed from independent draws is mathematically valid, it conflates differences between samples with differences between their corruptions. \teacher{} therefore samples one $(\sigma,\mathbf{n})$ per draw and shares it at inference and across the distillation batch. Replacing this shared draw with independent per-sample draws, while holding all other choices fixed, decreases $\mu$ by 9.2\%. Independent draws introduce unmatched, noise-dependent components into the gradients, causing the kernel to capture variation in the corruption draws rather than the relationship between the samples.

\paragraph{RQ3: Should gradient magnitudes be retained?}
\teacher{} normalizes each sample's gradient factors before constructing the kernel, thereby discarding the gradient magnitude that a literal reading of Eq.~\ref{eq:tau-kfac} would retain. Restoring this magnitude reduces $\mu$ by 6.2\%. At a given noise level, the gradient norm reflects how strongly the model responds to an individual sample, which depends on both sample difficulty and noise level. It is therefore a nuisance quantity for ranking training samples by their relationship to a query.

\paragraph{RQ4: Do the exact one-step Newton corrections help?}
The one-step Newton expansion in Eq.~\ref{eq:newton} contains two sample-specific factors omitted by \teacher{}: the training residual $r_i$ and the leverage $1-h_i$, which discounts samples already well accommodated by the fit. We find that introducing either correction produces no meaningful improvement: the changes in $\mu$ are both below 1\%. Thus, fidelity to the exact expansion does not appear to be a limiting factor, and the simpler kernel is preferable given its lower computational cost.

\paragraph{RQ5: Which curvature model should be used?}
\teacher{} preconditions its gradients using the Kronecker factors in Eq.~\ref{eq:kfac}, which are uncentered second moments of the observed activations and output gradients and thus constitute an empirical-Fisher approximation. In contrast,  K-FAC Influence constructs a Monte Carlo-sampled generalized Gauss--Newton (GGN) matrix by resampling the target of a squared loss. Replacing the empirical Fisher with the sampled GGN decreases $\mu$ by 5.1\%, which is smaller than the effect of changing the target in RQ1, indicating that methods' performance can be more affected by \emph{what} they differentiate than \emph{how} they approximate curvature.

\begin{table}[t]
\centering
\caption{Counterfactual AUCs for the ablation of the \teacher{} teacher on CIFAR-10. Each block isolates one design choice while holding the remaining components fixed. All variants use $M=32$.}
\label{tab:teacher-ablation}
\vspace{1.5mm}
\setlength{\tabcolsep}{6pt}
\resizebox{0.9\columnwidth}{!}{%
\newcommand{\indt}[1]{\hspace{4mm}}
\begin{tabular}{lcccccc}
\toprule
RQ/Variant & SSIM & SSCD & LPIPS & CLIP & $\mu$ & $\Delta$ \\
\midrule
\teacher{} (ours)
& 0.836 & 0.815 & 0.872 & 0.818 & 0.835 & --- \\
\midrule
RQ1: Attribution target & & & & & & \\
\indt{} $\lVert F_\theta\rVert_2^2$
 & 0.815 & 0.797 & 0.855 & 0.790 & 0.814 & $-$2.5\% \\
\indt{} $\omega(\sigma)\lVert\mathbf{x}-F_\theta\rVert_2^2$
 & 0.777 & 0.737 & 0.785 & 0.731 & 0.757 & $-$9.3\% \\
\midrule
RQ2: Corruption alignment & & & & & & \\
\indt{} Independent draws
 & 0.759 & 0.743 & 0.798 & 0.732 & 0.758 & $-$9.2\% \\
\midrule
RQ3: Gradient normalization & & & & & & \\
\indt{} $+$ Gradient magnitudes
 & 0.795 & 0.769 & 0.822 & 0.747 & 0.783 & $-$6.2\% \\
\midrule
RQ4: Newton corrections & & & & & & \\
\indt{} $+$ Residual $r_i$
 & 0.835 & 0.822 & 0.868 & 0.805 & 0.832 & $-$0.3\% \\
\indt{} $+$ Residual and leverage $r_i/(1-h_i)$
 & 0.842 & 0.815 & 0.882 & 0.813 & 0.838 & $+$0.3\% \\
\midrule
RQ5: Curvature model & & & & & & \\
\indt{} $+$ MC-sampled GGN
 & 0.792 & 0.773 & 0.838 & 0.767 & 0.792 & $-$5.1\% \\
\bottomrule
\end{tabular} }
\vspace{-10pt}
\end{table}

%% file: sections/conclusion.tex
\section{Conclusion}
\label{sec:conclusion}

We presented \teacher, a gradient-based TDA method for diffusion models that locally measures influence as the expected local discrepancy between the scores of the trained model and its leave-one-out counterfactual. Defined through the score, this target applies uniformly across diffusion variants, and a first-order expansion turns it into a Kronecker-factored gradient kernel that needs neither retraining, random projections, nor per-sample gradient storage. We further presented \student, which distills \teacher's rankings into embeddings computed from the diffusion model's own activations. With that, TDA reduces to cosine similarity on a low-dimensional space. Across three datasets, \teacher matches or surpasses unlearning-based attribution at a fraction of its cost, and \student retains most of this accuracy while attributing a query in milliseconds, faster than generating the sample itself. Our ablations also carry a broader lesson: the terms an exact derivation calls for, such as leverage corrections and schedule-dependent weights, consistently trade against ranking accuracy, and the simple aligned, normalized, uniformly aggregated kernel is the strongest practical choice.

\paragraph{Limitations and future work.}
We can enumerate three limitations of our work: (i) a general theory of when forward activations suffice to represent gradient-based influence is open; (ii) the distilled student is specific to the model and dataset it is trained on; (iii) our evaluation covers a single diffusion variant, although the formulation extends to DDPM, EDM, and flow matching. Besides limitations, we can also point to three directions for future work: (i) a distillation framework that transfers attribution knowledge across backbones, datasets, and teacher methods; (ii) stronger distribution-aware teachers, validated on the other diffusion variants; and (iii) reusing gradient information already computed during pre-training, so that teacher supervision and the student come at almost no additional cost.

%% file: sections/appendix.tex
\input{sections/related_work}

\input{sections/appendix/method_overview}

\input{sections/appendix/method_algorithm}

\input{sections/appendix/method_generality}

\input{sections/appendix/method_kronecker}

\input{sections/appendix/experiments}

%% file: sections/related_work.tex
\section{Related Work}
\label{sec:related-work}

\paragraph{Training data attribution.}
Training data attribution quantifies how training examples affect a target model behavior~\citep{hammoudeh2024training,deng2025survey}. Major approaches include influence functions, which approximate leave-one-out effects using local parameter perturbations~\citep{koh2017understanding}, weighted marginal-contribution methods such as Data Shapley and Data Banzhaf~\citep{ghorbani2019data,jia2019towards,wang2023data}, training-dynamics methods such as SGD-influence and TracIn~\citep{hara2019data,pruthi2020estimating}, and simulators such as Datamodels, which learn model behavior from repeated subset retraining~\citep{ilyas2022datamodels}. TRAK combines ideas from influence functions and datamodels with random gradient projections to improve scalability in large, non-convex models~\citep{park2023trak}. These approaches expose a recurring trade-off: retraining-based methods provide direct counterfactual semantics but require many model fits, whereas gradient-based approximations reduce retraining at the cost of gradient computation, curvature estimation, and storage.

\paragraph{Attribution for diffusion models.}
Earlier work extends attribution to generative models including GANs and VAEs~\citep{terashita2021influence,kong2021understanding}. For diffusion models, ensemble-based methods enable explicit data ablation by training models on structured subsets~\citep{dai2023training}, while gradient-based methods adapt TRAK or influence functions to the denoising process. Among the latter, Journey-TRAK attributes intermediate denoising states~\citep{georgiev2023journey}, D-TRAK studies alternative diffusion-specific gradient targets~\citep{zheng2024intriguing}, Diffusion-ReTrac corrects timestep-induced bias~\citep{xie2024data}, and DAS derives a diffusion-specific score with gradient normalization and KL-based motivation~\citep{lin2025diffusion}. Most closely related to \teacher{}, \citet{mlodozeniec2025influence} use K-FAC curvature approximations to estimate influence on proxies for generation probability, including diffusion loss, ELBO, and trajectory probability. In contrast, \teacher{} instead defines influence directly as the expected score discrepancy at noised states of the query. A complementary line of work uses mirrored unlearning, estimating influence by unlearning the generated query and measuring the resulting changes on training samples~\citep{ko2024mirrored,wang2024data,serra2026training}. These methods, however, require a separate optimization procedure for each query and cannot cache relevant information to reduce inference costs.

\paragraph{Efficient and representation-based attribution.}
Representation similarity supports efficient attribution through a single query encoding followed by nearest-neighbor retrieval, but generic visual similarity does not necessarily capture counterfactual influence~\citep{georgiev2023journey,wang2024data}. Prior work on representation-based attribution tunes visual representations using attribution-by-customization benchmarks~\citep{wang2023evaluating}, learns attribution models from representation changes during diffusion fine-tuning~\citep{brokman2024montrage}, or develops nonparametric attribution from representation geometry~\citep{zhao2025nonparametric}. Most closely related to \student{}, Fast Data Attribution distills an unlearning-based teacher into a standalone visual encoder using an offline collection of annotated pairs~\citep{wang2026fast}. In contrast to this, \student{} generates dense \teacher{} supervision online within each training batch and embeds the diffusion model's own internal activations. It directly distills per-query rankings rather than raw influence values, producing compact embeddings that retain model-specific attribution information while enabling efficient nearest-neighbor retrieval.

\subsection{Details of Gradient-based Methods}
Most gradient-based attribution methods for diffusion models, including ours, follow a common template: differentiate a scalar function of the model evaluated on a corrupted training sample, incorporate a curvature approximation, and compare each training sample with the query through a curvature-aware gradient inner product. As mentioned in Sec.~\ref{sec:method-teacher}, gradient-based methods differ primarily in the scalar target being differentiated, the curvature approximation, and the compression and normalization applied to the per-sample gradients. Table~\ref{tab:grad-landscape} organizes existing methods and the ablations in Table~\ref{tab:teacher-ablation} along these design axes.

\begin{table}[t]
\centering
\caption{Design choices of gradient-based diffusion attribution methods: D-TRAK~\citep{zheng2024intriguing}, DAS~\citep{lin2025diffusion}, K-FAC influence~\citep{mlodozeniec2025influence}, and \teacher{}. Here, $F_\theta$ denotes the denoiser output evaluated on a corrupted sample and $\mathbf{x}$ denotes the corresponding clean sample. Curvature specifies the preconditioner; kernel specifies how per-sample gradients are represented before computing inner products; and normalization specifies which representation, if any, is normalized at each timestep. Gradient normalization rescales each projected gradient to unit norm, whereas factor normalization rescales the two Kronecker factors independently, equivalently normalizing the corresponding rank-one gradient matrix by its Frobenius norm. Without normalization, large-gradient samples can dominate the aggregation across timesteps. The lower block varies individual components of \teacher{} and is evaluated quantitatively in Table~\ref{tab:teacher-ablation}.}
\label{tab:grad-landscape}
\vspace{1.5mm}
\footnotesize
\setlength{\tabcolsep}{4pt}
\begin{tabular}{lllll}
\toprule
Method & Target & Curvature & Kernel & Normalization \\
\midrule
D-TRAK
    & $\lVert F_\theta\rVert_2^2$
    & GN $+$ ridge
    & random projection
    & none \\
DAS
    & $\omega(\sigma)\lVert\mathbf{x}-F_\theta\rVert_2^2$
    & GN $+$ ridge
    & random projection
    & gradient \\
K-FAC influence
    & $\omega(\sigma)\lVert\mathbf{x}-F_\theta\rVert_2^2$
    & MC-GGN
    & K-FAC
    & none \\
\teacher{} (ours)
    & $\langle\mathbf{1},F_\theta\rangle$
    & empirical Fisher
    & K-FAC
    & factors \\
\midrule
\multicolumn{5}{l}{\emph{Ablation variants of \teacher{}}} \\
\quad RQ1a: squared output
    & $\lVert F_\theta\rVert_2^2$
    & empirical Fisher
    & K-FAC
    & factors \\
\quad RQ1b: denoising loss
    & $\omega(\sigma)\lVert\mathbf{x}-F_\theta\rVert_2^2$
    & empirical Fisher
    & K-FAC
    & factors \\
\quad RQ2: independent draws
    & $\langle\mathbf{1},F_\theta\rangle$
    & empirical Fisher
    & K-FAC
    & factors \\
\quad RQ3: gradient magnitude
    & $\langle\mathbf{1},F_\theta\rangle$
    & empirical Fisher
    & K-FAC
    & \textbf{none} \\
\quad RQ4a: $+\,r_i$
    & $\langle\mathbf{1},F_\theta\rangle$
    & empirical Fisher
    & K-FAC
    & factors \\
\quad RQ4b: $+\,r_i/(1-h_i)^2$
    & $\langle\mathbf{1},F_\theta\rangle$
    & empirical Fisher
    & K-FAC
    & factors \\
\quad RQ5: sampled curvature
    & $\langle\mathbf{1},F_\theta\rangle$
    & \textbf{MC-GGN}
    & K-FAC
    & factors \\
\bottomrule
\end{tabular}
\end{table}

%% file: sections/appendix/method_overview.tex
\section{An Overview of \teacher and \student}
\label{app:overview}
\begin{figure}[t]
\centering
\includegraphics[width=\linewidth]{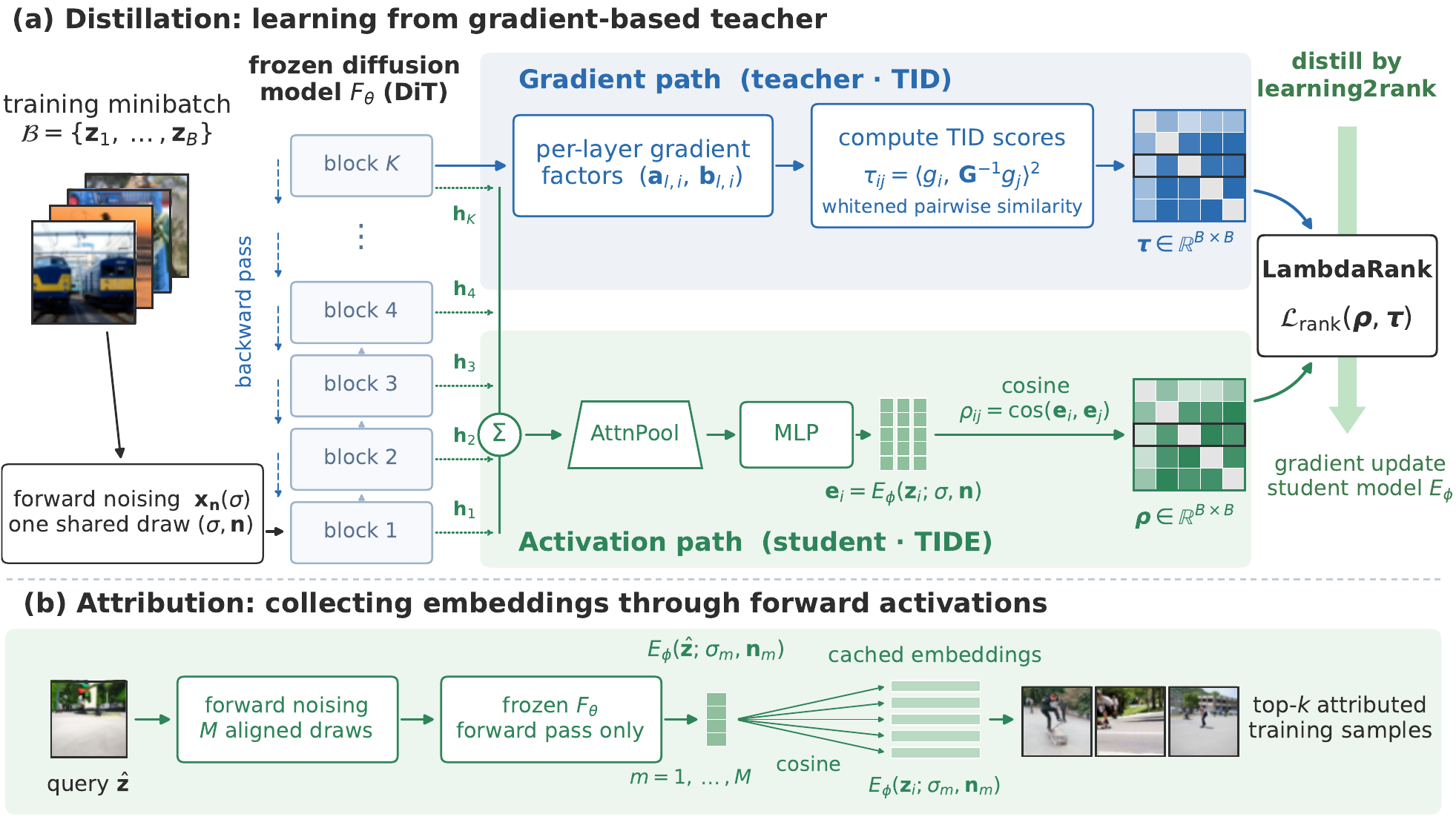}
\vspace{-10pt}
\caption{Overview of \teacher{} and \student{}. }
\label{fig:method}
\end{figure}

Figure~\ref{fig:method} illustrates how the teacher of Sec.~\ref{sec:method-teacher} and the student of Sec.~\ref{sec:method-student} share a single pass through the frozen diffusion model. Panel~(a) shows one distillation step of Algorithm~\ref{alg:distill}. A minibatch is noised with one shared draw $(\sigma,\mathbf{n})$ under the aligned regime, and passed through $F_\theta$ once. Along the gradient path, the backward pass yields the per-layer factors $(\mathbf{a}_{l,i},\mathbf{b}_{l,i})$, which the Kronecker-factored kernel of Eq.~\ref{eq:kfac} turns into the within-batch teacher matrix $\boldsymbol{\tau}$ of \teacher scores. Along the activation path, the residual streams $\mathbf{h}_1,\dots,\mathbf{h}_K$ of the same forward pass are summed, attention-pooled, and projected into the \student embeddings of Eq.~\ref{eq:embeddings}, whose cosine similarities form the student matrix $\boldsymbol{\rho}$. The student and teacher matrices are compared row by row by the ranking loss of Eq.~\ref{eq:lambdarank}, with each sample acting as anchor and the diagonal excluded. The teacher is detached, so only the embedding model $E_\phi$ is updated. Panel~(b) shows attribution at inference time with \student: a generated query is noised with the $M$ aligned draws, embedded from forward activations only, and scored by cosine similarity against training embeddings that are computed once and cached, so no backward pass is needed.

%% file: sections/appendix/method_algorithm.tex
\paragraph{Distillation Algorithm.}
Algorithm~\ref{alg:distill} gives the procedural form of the distillation step illustrated in Figure~\ref{fig:method}(a). The Kronecker factor inverses of Eq.~\ref{eq:kfac} are estimated once from the training set before distillation begins and are held constant, so the only quantities recomputed per batch are the per-sample gradient factors and embeddings. Each batch uses a single shared draw $(\sigma,\mathbf{n})$ under the aligned regime of Sec.~\ref{sec:method-teacher}, and one forward--backward pass per sample supplies the information for both paths: the residual-stream activations feed the embedding of Eq.~\ref{eq:embeddings} and the per-layer gradient factors feed the teacher kernel of Eq.~\ref{eq:kfac}. The teacher matrix $\boldsymbol{\tau}$ is detached, so the gradient step on the ranking loss of Eq.~\ref{eq:lambdarank} updates only the embedding parameters $\phi$ while the diffusion model $F_\theta$ is never modified. The per-batch cost is therefore that of one backward pass per sample plus $O(B^2)$ inner products, and no teacher scores are stored across batches.

\begin{algorithm}[t]
\caption{Batched pair-wise score distillation}
\label{alg:distill}
\begin{algorithmic}[1]
\small
\Require frozen diffusion model $F_\theta$; training set $\mathcal{Z}$; student embedder $E_\phi$; K-FAC factor inverses $\{\mathbf{A}_l^{-1},\mathbf{B}_l^{-1}\}$ estimated once and offline from $\mathcal{Z}$ (Sec.~\ref{sec:method-teacher}).
\For{each training batch $\mathcal{B}=\{\mathbf{z}_1,\dots,\mathbf{z}_B\}\subset\mathcal{Z}$}
\State draw one shared $(\sigma,\mathbf{n})$ with $\sigma$ from the sampler's schedule and $\mathbf{n}\sim\mathcal{N}(\mathbf{0},I)$
\For{each $\mathbf{z}_i=\{\mathbf{x},\mathbf{c}\}\in\mathcal{B}$}
\State forward--backward pass of $F_\theta\big(\mathbf{x}_{\sigma,\mathbf{n}};\sigma,\mathbf{c}\big)$
\State collect the activations $\{\mathbf{h}_{k}(\mathbf{z}_i;\sigma,\mathbf{n})\}_{k=1}^{K}$ and compute embeddings $\mathbf{e}_i$ using Eq.~\ref{eq:embeddings}
\State collect the gradient factors $(\mathbf{a}_{l,i},\mathbf{b}_{l,i})$
\EndFor
\For{all pairs $(i,j)$ in $\mathcal{B}$}
\State compute teacher scores $\tau_{ij}$ (detached) using gradient factors $\{(\mathbf{a}_{l,i},\mathbf{b}_{l,i}),(\mathbf{a}_{l,j},\mathbf{b}_{l,j})\}$ and Eq.~\ref{eq:kfac}
\State compute student scores $\rho_{ij}$ using embeddings $(\mathbf{e}_i,\mathbf{e}_j)$ and cosine similarity
\EndFor
\State update $\phi$ by one gradient step on $\mathcal{L}_{\mathrm{rank}}(\rho,\tau)$ using Eq.~\ref{eq:lambdarank}
\EndFor
\State \Return $E_\phi$
\end{algorithmic}
\end{algorithm}

%% file: sections/appendix/method_generality.tex
\section{Handling Multiple Diffusion Variants}
\label{app:method-proof}

\subsection{Theoretical Background}
\label{app:theoret-bckgrd}

We briefly recall the SDE view of diffusion models~\citep{song2020score}, indexed directly by the noise scale $\sigma$ used throughout the paper. We first fix the notation of the expectations used below, which we keep out of the main text. The expectation $\mathbb{E}_{\mathbf{z},\sigma,\mathbf{n}}$ is over a training instances $\mathbf{z}=\{\mathbf{x},\mathbf{c}\}$ drawn from $\mathcal{Z}$, noise scales $\sigma$ drawn from the variant's training noise-scale distribution, and noise vectors $\mathbf{n}\sim\mathcal{N}(\mathbf{0},I)$. In $\mathbb{E}_{\sigma,\mathbf{n}}$ we omit the draw of $\mathbf{z}$. The network $F_\theta(\mathbf{x}_{\sigma,\mathbf{n}};\sigma,\mathbf{c})$ is evaluated on the forward-noised item $\mathbf{x}_{\sigma,\mathbf{n}}$, defined below, with the noise scale $\sigma$ and the conditioning $\mathbf{c}$ as additional inputs. All densities in this section are conditioned on $\mathbf{c}$, which we suppress when unambiguous.

\paragraph{Forward and reverse processes.}
Given a training instance $\mathbf{z}=\{\mathbf{x},\mathbf{c}\}$, a diffusion model specifies a stochastic process $\{\mathbf{x}_\sigma\}_{\sigma\in[0,\sigma_{\max}]}$ that starts at the data item, $\mathbf{x}_\sigma=\mathbf{x}$ at $\sigma=0$, and ends at a tractable prior at $\sigma_{\max}$, by solving the It\^{o} SDE
\[
d\mathbf{x}_\sigma=\mathbf{f}(\mathbf{x}_\sigma,\sigma)\,d\sigma+\eta(\sigma)\,d\mathbf{W}_\sigma,
\]
where $\mathbf{f}(\cdot,\sigma):\mathbb{R}^d\to\mathbb{R}^d$ is the drift coefficient, $\eta:\mathbb{R}\to\mathbb{R}$ the scalar diffusion coefficient, and $\mathbf{W}_\sigma$ a standard Wiener process. We write $q(\mathbf{x}_\sigma\mid\mathbf{x})$ for the transition kernel of this process from the clean item $\mathbf{x}$ to noise scale $\sigma$, and $p_\sigma$ for the resulting marginal of $\mathbf{x}_\sigma$. \citet{anderson1982reverse} shows that the reversal of this process is itself a diffusion
\[
d\mathbf{x}_\sigma=\big[\mathbf{f}(\mathbf{x}_\sigma,\sigma)-\eta(\sigma)^2\,\nabla_{\mathbf{x}_\sigma}\log p_\sigma(\mathbf{x}_\sigma)\big]\,d\sigma+\eta(\sigma)\,d\bar{\mathbf{W}}_\sigma,
\]
where $\bar{\mathbf{W}}_\sigma$ is a standard Wiener process when $\sigma$ runs backwards from $\sigma_{\max}$ to $0$ and $d\sigma$ is an infinitesimal negative step. Generation thus reduces to knowing the score $\nabla_{\mathbf{x}_\sigma}\log p_\sigma(\mathbf{x}_\sigma)$ at every noise scale. The samplers we consider discretize this reverse process, or its deterministic probability-flow counterpart~\citep{song2020score}, on the schedule $\sigma_{\max}=\sigma_T>\dots>\sigma_1=\sigma_{\min}$ of Sec.~\ref{sec:method}.

\paragraph{Score matching.}
The score $\nabla_{\mathbf{x}_\sigma}\log p_\sigma(\mathbf{x}_\sigma)$ is learned by a noise-conditioned network $s_\theta$ via denoising score matching,
\[
\min_\theta\;\mathbb{E}_{\mathbf{z},\sigma,\mathbf{n}}\Big[\lambda(\sigma)\,\big\lVert s_\theta(\mathbf{x}_{\sigma,\mathbf{n}};\sigma,\mathbf{c})-\nabla_{\mathbf{x}_\sigma}\log q(\mathbf{x}_\sigma\mid\mathbf{x})\big|_{\mathbf{x}_\sigma=\mathbf{x}_{\sigma,\mathbf{n}}}\big\rVert_2^2\Big],
\]
where $\lambda:[\sigma_{\min},\sigma_{\max}]\to\mathbb{R}_{>0}$ is a weighting function and $\mathbf{x}_{\sigma,\mathbf{n}}\sim q(\cdot\mid\mathbf{x})$. For any positive $\lambda$, the minimizer equals the marginal score almost everywhere. The objective only requires the transition kernel $q$. When the drift is affine and has the scalar linear form $\mathbf{f}(\mathbf{x}_\sigma,\sigma)=f(\sigma)\mathbf{x}_\sigma$, as in every variant we consider, the transition kernel is Gaussian,
\[
q(\mathbf{x}_\sigma\mid\mathbf{x})
=\mathcal{N}\bigl(\mu(\sigma)\mathbf{x},\nu(\sigma)^2I\bigr),
\]
with closed-form signal and noise coefficients
\[
\mu(\sigma)=\exp\!\left(\int_0^\sigma f(u)\,du\right)
\qquad\text{and}\qquad
\nu(\sigma)^2=\mu(\sigma)^2
\int_0^\sigma\frac{\eta(u)^2}{\mu(u)^2}\,du,
\]
so that a forward-noised state and the corresponding score-matching target can be written directly in terms of the noise draw $\mathbf{n}$. Using the notation of Sec.~\ref{sec:method} the forward-noised state is
\[
\mathbf{x}_{\sigma,\mathbf{n}}=\mu(\sigma)\,\mathbf{x}+\nu(\sigma)\,\mathbf{n},
\]
and the score-matching target is
\[
\nabla_{\mathbf{x}_\sigma}\log q(\mathbf{x}_\sigma\mid\mathbf{x})\Big|_{\mathbf{x}_\sigma=\mathbf{x}_{\sigma,\mathbf{n}}}=-\frac{\mathbf{x}_{\sigma,\mathbf{n}}-\mu(\sigma)\,\mathbf{x}}{\nu(\sigma)^2}=-\frac{\mathbf{n}}{\nu(\sigma)}.
\]

For the variance-exploding SDE, $\mathbf{f}(\mathbf{x}_\sigma,\sigma)=\mathbf{0}$ and $\eta(\sigma)=\sqrt{2\sigma}$ gives $(\mu,\nu)=(1,\sigma)$, and this becomes the process underlying EDM~\citep{karras2022elucidating}. For the variance-preserving SDE, $\mathbf{f}(\mathbf{x}_\sigma,\sigma)=-\tfrac12\beta(\sigma)\,\mathbf{x}_\sigma$ and $\eta(\sigma)=\sqrt{\beta(\sigma)}$ for a positive noise schedule $\beta$ gives $(\mu,\nu)=(\sqrt{\alpha(\sigma)},\sqrt{1-\alpha(\sigma)})$ with $\alpha(\sigma)=\exp\big(-\int_0^\sigma\beta(s)\,ds\big)$, and this becomes the continuous limit of DDPM~\citep{ho2020denoising}. The linear interpolant of flow matching, $(\mu,\nu)=(1-\sigma,\sigma)$ on $\sigma\in(0,1)$, is usually presented as an ODE rather than an SDE, but has the same Gaussian conditional kernel, so its conditional score is given by the same expression.

\paragraph{From network output to score.}
In practice, the network $F_\theta$ is not trained to output the score itself but a schedule-dependent regression target, such as the clean instance, the noise, or a velocity. The network is evaluated on the forward-noised instance as $F_\theta(\mathbf{x}_{\sigma,\mathbf{n}};\sigma,\mathbf{c})$, with the noise scale $\sigma$ and the conditioning $\mathbf{c}$ as additional inputs, and $\hat{\mathbf{x}}_{\sigma,\mathbf{n}}$ denotes the forward noising of an item $\mathbf{x}$ with the same draw $(\sigma,\mathbf{n})$. In this notation, the training loss $L$ implements the diffusion variant together with its forward noising.

For DDPM~\citep{ho2020denoising}, $F_\theta$ predicts the noise,
\begin{equation*}
L(F_\theta)=\mathbb{E}_{\mathbf{z},\sigma,\mathbf{n}}\Big[\big\lVert\mathbf{n}-F_\theta\big(\mathbf{x}_{\sigma,\mathbf{n}};\sigma,\mathbf{c}\big)\big\rVert_2^{2}\Big]\quad\text{with}\quad\mathbf{x}_{\sigma,\mathbf{n}}=\sqrt{\alpha(\sigma)}\,\mathbf{x}+\sqrt{1-\alpha(\sigma)}\,\mathbf{n}.
\end{equation*}
For EDM~\citep{karras2022elucidating}, $F_\theta$ predicts the clean instance under a noise-scale-dependent weight $\omega$,
\begin{equation*}
L(F_\theta)=\mathbb{E}_{\mathbf{z},\sigma,\mathbf{n}}\Big[\omega(\sigma)\,\big\lVert\mathbf{x}-F_\theta\big(\mathbf{x}_{\sigma,\mathbf{n}};\sigma,\mathbf{c}\big)\big\rVert_2^{2}\Big]\quad\text{with}\quad\mathbf{x}_{\sigma,\mathbf{n}}=\mathbf{x}+\sigma\mathbf{n}.
\end{equation*}
For flow matching~\citep{lipman2022flow,liu2022flow,albergo2022stochastic}, $F_\theta$ regresses the conditional velocity $\mathbf{n}-\mathbf{x}$ along the linear path, with $\sigma\sim\mathcal{U}(0,1)$,
\begin{equation*}
L(F_\theta)=\mathbb{E}_{\mathbf{z},\sigma,\mathbf{n}}\Big[\big\lVert\mathbf{n}-\mathbf{x}-F_\theta\big(\mathbf{x}_{\sigma,\mathbf{n}};\sigma,\mathbf{c}\big)\big\rVert_2^{2}\Big]\quad\text{with}\quad\mathbf{x}_{\sigma,\mathbf{n}}=(1-\sigma)\,\mathbf{x}+\sigma\,\mathbf{n}.
\end{equation*}
Each target determines the noise $\mathbf{n}$ as a function of $F_\theta$ and the noisy state $\mathbf{x}_{\sigma,\mathbf{n}}$, so the score estimate $s_\theta$ is recovered by substitution into the conditional score $-\mathbf{n}/\nu(\sigma)$ derived above, and for all common parameterizations the resulting map is affine.

\begin{lemma}[Affine output-to-score map; \citealp{ho2020denoising,karras2022elucidating}]
\label{lem:m-affine}
For the common diffusion parameterizations, the score estimate is an affine function of the network output with coefficients depending only on the noise schedule,
\[
s_\theta=\kappa(\sigma)F_\theta+b(\sigma)\mathbf{x}.
\]
For the standard DDPM $\epsilon$-prediction parameterization,
\[
s_\theta=-\frac{F_\theta}{\sqrt{1-\alpha(\sigma)}},\qquad
\kappa(\sigma)=-\frac1{\sqrt{1-\alpha(\sigma)}},\qquad
b(\sigma)=0,
\]
while for EDM,
\[
s_\theta=\frac{F_\theta-\mathbf{x}}{\sigma^2},\qquad
\kappa(\sigma)=\frac1{\sigma^2},\qquad
b(\sigma)=-\frac1{\sigma^2},
\]
and for flow matching on the linear path $\mathbf{x}_{\sigma,\mathbf{n}}=(1-\sigma)\mathbf{x}+\sigma\mathbf{n}$, where $F_\theta$ predicts the velocity $\mathbf{n}-\mathbf{x}$ so that $\mathbf{n}=\mathbf{x}_{\sigma,\mathbf{n}}+(1-\sigma)F_\theta$,
\[
s_\theta=-\frac{(1-\sigma)\,F_\theta+\mathbf{x}_{\sigma,\mathbf{n}}}{\sigma},\qquad \kappa(\sigma)=-\frac{1-\sigma}{\sigma},\qquad b(\sigma)=-\frac1{\sigma}.
\]
\end{lemma}

\subsection{Variant Weights}
\label{app:method-weights}

This section instantiates the conversion weight $w_\sigma=\kappa(\sigma)^2$ of Eq.~\ref{eq:tau-output} for the common parameterizations. By Lemma~\ref{lem:m-affine}, the score estimate is affine in the network output, $s_\theta=\kappa(\sigma)F_\theta+b(\sigma)\mathbf{x}$. Since the term $b(\sigma)\mathbf{x}$ is shared by any two models evaluated at the same state, it cancels when computing score differences, giving
\[
\lVert s_\theta-s_{\theta^{\setminus i}}\rVert_2^2
=\kappa(\sigma)^2\,\lVert F_\theta-F_{\theta^{\setminus i}}\rVert_2^2
\]
and hence we obtain Eq.~\ref{eq:tau-output} with $w_\sigma=\kappa(\sigma)^2$. The coefficients follow from expressing the conditional score $\nabla_{\mathbf{x}_{\mathbf{n}}}\log q(\mathbf{x}_{\mathbf{n}}\,|\,\mathbf{x})$ in each variant's output parameterization.
\begin{itemize}
\item EDM: $F_\theta$ predicts the clean item and $s_\theta=(F_\theta-\mathbf{x}_{\mathbf{n}})/\sigma^2$, so
\[
\kappa(\sigma)=\frac1{\sigma^2}\qquad\text{and}\qquad w_\sigma=\frac1{\sigma^4}.
\]

\item DDPM ($\epsilon$-prediction): $s_\theta=-F_\theta/\sqrt{1-\alpha(\sigma)}$, so
\[
\kappa(\sigma)=-\frac1{\sqrt{1-\alpha(\sigma)}}\qquad\text{and}\qquad w_\sigma=\frac1{1-\alpha(\sigma)}.
\]

\item $\mathbf{v}$-prediction: With $\mathbf{x}_{\mathbf{n}}=\sqrt{\alpha}\,\mathbf{x}+\sqrt{1-\alpha}\,\mathbf{n}$ and $F_\theta$ predicting $\mathbf{v}=\sqrt{\alpha}\,\mathbf{n}-\sqrt{1-\alpha}\,\mathbf{x}$, the identities $\mathbf{n}=\sqrt{1-\alpha}\,\mathbf{x}_{\mathbf{n}}+\sqrt{\alpha}\,F_\theta$ and $s_\theta=-\mathbf{n}/\sqrt{1-\alpha}$ give
\[
\kappa(\sigma)=-\frac{\sqrt{\alpha(\sigma)}}{\sqrt{1-\alpha(\sigma)}}\qquad\text{and}\qquad
w_\sigma=\frac{\alpha(\sigma)}{1-\alpha(\sigma)},
\]
the signal-to-noise ratio.

\item Flow matching: For the linear path $\mathbf{x}_{\mathbf{n}}(\sigma)=(1-\sigma)\mathbf{x}+\sigma\mathbf{n}$, $\sigma\in(0,1)$, with velocity output $F_\theta$ regressing $\mathbf{n}-\mathbf{x}$, eliminating $\mathbf{x}$ via $\mathbf{x}=\mathbf{x}_{\mathbf{n}}-\sigma F_\theta$ gives $\mathbf{n}=\mathbf{x}_{\mathbf{n}}+(1-\sigma)F_\theta$; the conditional score of the path's Gaussian kernel then yields
\[
\kappa(\sigma)=-\frac{1-\sigma}{\sigma}\qquad\text{and}\qquad
w_\sigma=\frac{(1-\sigma)^2}{\sigma^2},
\]
up to the path's noise normalization.
\end{itemize}

In all cases, $w_\sigma$ depends only on the noise schedule and is bounded above and below by positive constants on the truncated window $[\sigma_{\min},\sigma_{\max}]$ with $\sigma_{\min}>0$ (and $\sigma_{\max}<1$ for flow matching).

%% file: sections/appendix/method_kronecker.tex
\section{Factorized Gradients Beyond Linear Layers}
\label{app:layer-extensions}

The main text presents the K-FAC decomposition for a linear layer without a token dimension. Here, we give the corresponding formulas for the other parameterized layers commonly used in diffusion architectures. Throughout this section, $\mathbf{a}_{l,i,t}\in\mathbb{R}^{N_{l,\mathrm{in}}}$ and $\mathbf{b}_{l,i,t}\in\mathbb{R}^{N_{l,\mathrm{out}}}$ denote the two gradient factors for training sample $\mathbf{z}_i$ at position $t$. The position index may represent a sequence token or a spatial location. The per-example layer gradient has the general form
\begin{equation}
\label{eq:general-factorized-gradient}
g_l(\mathbf{z}_i)=\sum_{t=1}^{T_l}\mathbf{b}_{l,i,t}\mathbf{a}_{l,i,t}^{\top}.
\end{equation}
Consequently, the inner product between two layer gradients can be evaluated without materializing either gradient:
\begin{equation}
\label{eq:general-ghost-inner-product}
\big\langle g_l(\mathbf{z}_i),g_l(\mathbf{z}_j)\big\rangle
=\sum_{t=1}^{T_l}\sum_{s=1}^{T_l}
\big(\mathbf{a}_{l,i,t}^{\top}\mathbf{a}_{l,j,s}\big)
\big(\mathbf{b}_{l,i,t}^{\top}\mathbf{b}_{l,j,s}\big).
\end{equation}
The corresponding squared norm is obtained by setting $i=j$. Under K-FAC, positions are treated as observations, giving
\begin{equation}
\label{eq:general-kfac-factors}
\mathbf{A}_l=\frac{1}{BT_l}\sum_{i,t}\mathbf{a}_{l,i,t}\mathbf{a}_{l,i,t}^{\top},
\qquad
\mathbf{B}_l=\frac{1}{BT_l}\sum_{i,t}\mathbf{b}_{l,i,t}\mathbf{b}_{l,i,t}^{\top}.
\end{equation}
The preconditioned inner product therefore remains factorized:
\begin{equation}
\label{eq:general-preconditioned-ghost-inner-product}
\left\langle g_l(\mathbf{z}_i),\mathbf{G}_l^{-1}g_l(\mathbf{z}_j)\right\rangle
\approx\sum_{t=1}^{T_l}\sum_{s=1}^{T_l}
\big(\mathbf{a}_{l,i,t}^{\top}\mathbf{A}_l^{-1}\mathbf{a}_{l,j,s}\big)
\big(\mathbf{b}_{l,i,t}^{\top}\mathbf{B}_l^{-1}\mathbf{b}_{l,j,s}\big).
\end{equation}

\paragraph{Linear layers with token dimensions.}
For a linear transformation applied independently to $T_l$ tokens, $\mathbf{a}_{l,i,t}$ is the input activation and $\mathbf{b}_{l,i,t}$ is the gradient with respect to the pre-activation output. Its weight gradient is exactly Eq.~\ref{eq:general-factorized-gradient}. When no token dimension is present, $T_l=1$, and Eq.~\ref{eq:general-preconditioned-ghost-inner-product} reduces to Eq.~\ref{eq:kfac} in the main text. A bias can be included by augmenting every activation factor as $\tilde{\mathbf{a}}_{l,i,t}=[\mathbf{a}_{l,i,t}^{\top},1]^{\top}$.

\paragraph{Convolutional layers.}
Consider a two-dimensional convolution with kernel size $K_h\times K_w$ and $C_{\mathrm{in}}$ input channels. After applying the standard \emph{unfold} (im2col) operator, let
\begin{equation}
\mathbf{a}_{l,i,t}\in\mathbb{R}^{C_{\mathrm{in}}K_hK_w},
\qquad
\mathbf{b}_{l,i,t}\in\mathbb{R}^{C_{\mathrm{out}}},
\qquad
T_l=H_{\mathrm{out}}W_{\mathrm{out}},
\end{equation}
where $\mathbf{a}_{l,i,t}$ is the input patch associated with output location $t$ and $\mathbf{b}_{l,i,t}$ is the corresponding pre-activation output gradient. The flattened kernel gradient is
\begin{equation}
g_l(\mathbf{z}_i)=\sum_{t=1}^{T_l}\mathbf{b}_{l,i,t}\mathbf{a}_{l,i,t}^{\top},
\end{equation}
so Eqs.~\ref{eq:general-ghost-inner-product}--\ref{eq:general-preconditioned-ghost-inner-product} apply directly. The same construction extends to one- and three-dimensional convolutions by replacing the receptive-field dimension accordingly. Convolutional biases are again incorporated by appending a constant to each unfolded input patch.

\paragraph{Transposed convolutional layers.}
For a transposed convolution, the roles of the unfolded quantities are reversed. Let $t$ index the $T_l$ input spatial locations, let $\mathbf{a}_{l,i,t}\in\mathbb{R}^{C_{\mathrm{in}}}$ be the input at location $t$, and let $\mathbf{b}_{l,i,t}\in\mathbb{R}^{C_{\mathrm{out}}K_hK_w}$ be the corresponding kernel-shaped patch extracted from the output gradient. Up to the fixed permutation induced by the kernel layout,
\begin{equation}
g_l(\mathbf{z}_i)=\sum_{t=1}^{T_l}\mathbf{a}_{l,i,t}\mathbf{b}_{l,i,t}^{\top}.
\end{equation}
Because transposition does not change a Frobenius inner product, the same ghost-inner-product and K-FAC formulas apply after exchanging the two factors. Unlike an ordinary convolution, the bias dimension $C_{\mathrm{out}}$ does not match the unfolded output-gradient dimension $C_{\mathrm{out}}K_hK_w$; we therefore handle the bias as a separate parameter block, with per-example gradient
\begin{equation}
g_{l,\mathrm{bias}}(\mathbf{z}_i)=\sum_{u}\frac{\partial\mathcal{L}_i}{\partial \mathbf{y}_{l,i,u}}.
\end{equation}
This avoids introducing padded factors solely to combine the weight and bias blocks.

\paragraph{Normalization layers.}
For LayerNorm with scale $\boldsymbol{\gamma}_l$ and shift $\boldsymbol{\beta}_l$,
\begin{equation}
\mathbf{y}_{l,i,t}=\boldsymbol{\gamma}_l\odot\hat{\mathbf{x}}_{l,i,t}+\boldsymbol{\beta}_l,
\end{equation}
where $\hat{\mathbf{x}}_{l,i,t}$ is the normalized input. Writing $\mathbf{b}_{l,i,t}=\partial\mathcal{L}_i/\partial\mathbf{y}_{l,i,t}$, the per-example parameter gradients are
\begin{equation}
g_{l,\gamma}(\mathbf{z}_i)=\sum_t\hat{\mathbf{x}}_{l,i,t}\odot\mathbf{b}_{l,i,t},
\qquad
g_{l,\beta}(\mathbf{z}_i)=\sum_t\mathbf{b}_{l,i,t}.
\end{equation}
Thus, their exact inner product is
\begin{equation}
\label{eq:norm-inner-product}
\big\langle g_l(\mathbf{z}_i),g_l(\mathbf{z}_j)\big\rangle
=g_{l,\gamma}(\mathbf{z}_i)^{\top}g_{l,\gamma}(\mathbf{z}_j)
+g_{l,\beta}(\mathbf{z}_i)^{\top}g_{l,\beta}(\mathbf{z}_j).
\end{equation}
These gradients are vectors rather than outer products, so we use a diagonal curvature approximation for normalization parameters. The same elementwise formulation applies to other affine normalization layers, including GroupNorm and InstanceNorm, after summing over the axes on which their scale and shift parameters are shared.

Together, these constructions cover the principal parameterized layer types in the diffusion architectures used in our experiments while preserving factorized computation wherever an outer-product structure is available.

%% file: sections/appendix/experiments.tex
\section{Experimental Details}

\label{app:exp-details}

\paragraph{Models, training, and compute.}
All experiments are run on a single NVIDIA H100 80\,GB GPU. Table~\ref{tab:setup} summarizes the datasets, model architectures, training configurations, and evaluation protocol, while Table~\ref{tab:attrib-settings} reports the settings of \teacher{} and \student{}. All diffusion models use DiT backbones trained under the EDM formulation, following the code released with MUCS~\citep{serra2026training}. We use an exponential moving average of the diffusion model parameters during training. Each counterfactual model is retrained with the same configuration and random seed as its full-data counterpart, excluding only the selected influential samples. This paired design reduces variation due to optimization and isolates the effect of removing the attributed data.

\begin{table}[t]
\centering
\caption{Dataset, model, training, sampling, and evaluation configurations. Attribution-specific settings are reported separately in Table~\ref{tab:attrib-settings}.}
\label{tab:setup}
\vspace{1.5mm}
\small
\setlength{\tabcolsep}{4pt}
\begin{tabular}{llccc}
\toprule
& & CIFAR-10 & ArtBench-10 & MS-COCO \\
\midrule
\multirow{3}{*}{\rotatebox[origin=c]{90}{Data}}
& Training samples $N$     & 50{,}000 & 49{,}917 & 118{,}287 \\
& Resolution               & $32{\times}32$ & $64{\times}64$ & $64{\times}64$ \\
& Conditioning             & none & class label & CLIP text embedding \\
\midrule
\multirow{8}{*}{\rotatebox[origin=c]{90}{Backbone}}
& Architecture             & DiT & DiT & DiT \\
& Transformer blocks       & 12 & 12 & 12 \\
& Model width              & 768 & 768 & 768 \\
& Attention heads          & 12 & 12 & 12 \\
& Patch size               & $2{\times}2$ & $2{\times}2$ & $2{\times}2$ \\
& Stem channels            & $\{4,4\}$ & $\{16,16\}$ & $\{16,16\}$ \\
& Stem resampling          & $\{1,1\}$ & $\{1,2\}$ & $\{1,2\}$ \\
& Conditioning width       & 256 & 256 & 1{,}024 \\
\midrule
\multirow{7}{*}{\rotatebox[origin=c]{90}{Optimization}}
& Optimizer                & AdamW & AdamW & AdamW \\
& Learning rate            & $1{\times}10^{-4}$ & $2{\times}10^{-4}$ & $2{\times}10^{-4}$ \\
& Weight decay             & 0.01 & 0.01 & 0.01 \\
& Batch size               & 128 & 128 & 128 \\
& Updates per epoch        & 1{,}000 & 1{,}000 & 1{,}000 \\
& Epochs                   & 200 & 250 & 300 \\
& EMA decay                & 0.999 & 0.999 & 0.999 \\
\midrule
\multirow{4}{*}{\rotatebox[origin=c]{90}{Sampler}}
& Schedule                 & Karras & Karras & Karras \\
& Steps                    & 32 & 32 & 32 \\
& $\sigma$ range           & $[0.002,80]$ & $[0.002,80]$ & $[0.002,80]$ \\
& $\rho$                   & 7 & 7 & 7 \\
\midrule
\multirow{3}{*}{\rotatebox[origin=c]{90}{Protocol}}
& Queries per seed         & 20 & 20 & 20 \\
& Seeds                    & 6 & 6 & 6 \\
& Per-query budget         & $2\%$ & $2\%$ & $2\%$ \\
\bottomrule
\end{tabular}
\end{table}

\begin{table}[t]
\centering
\caption{Attribution settings for \teacher{} and \student{}. We use the same configuration on all three datasets. Table~\ref{tab:main} reports results with $M{=}32$, while Fig.~\ref{fig:efficiency} varies the budget over $\{1,32,100\}$.}
\label{tab:attrib-settings}
\small
\setlength{\tabcolsep}{6pt}
\begin{tabular}{llc}
\toprule
& & Setting \\
\midrule
\multirow{5}{*}{\rotatebox[origin=c]{90}{\teacher{}}}
& Timestep schedule           & Karras \\
& Corruption draw             & shared \\
& Relative damping $\alpha$   & 0.1 \\
& Prenormalization            & per sample \\
& Inference draws $M$               & 32 \\
& Inference draw alignment    & aligned $(\sigma,\mathbf{n})$ \\
\midrule
\multirow{13}{*}{\rotatebox[origin=c]{90}{\student{}}}
& Input features              & residual stream \\
& Residual blocks used        & 12 \\
& Residual combination        & sum \\
& Pooling                     & attention \\
& Hidden width                & 512 \\
& Embedding dimension         & 768 \\
& Similarity head             & cosine \\
& Scale $\alpha=e^{\rho}$, init.\ & 32 \\
& Distillation loss           & LambdaRank \\
& Epochs                      & 50 \\
& Batch size                  & 128 \\
& Learning rate               & $5{\times}10^{-5}$ \\
& Inference draws $M$         & 32 \\
& Inference draw alignment    & aligned $(\sigma,\mathbf{n})$ \\
\bottomrule
\end{tabular}
\end{table}

\paragraph{Protocol details.}
For each seed, we generate a fixed set of 20 held-out queries using the full-data model. Every TDA method selects the top $2\%$ of the training set for each query, after which we retrain the model without the union of the 20 selected sets. Because the per-query sets overlap, the number of distinct samples removed depends on the method. Rankings that repeatedly select the same samples yield a smaller union than rankings that are more query-specific. As shown in Table~\ref{tab:removal}, the resulting removal fractions range from $21\%$ to $33\%$. The random control produces the largest union because its per-query selections overlap the least, whereas the model-agnostic encoders tend to retrieve visually central samples shared across queries and therefore produce smaller unions. Appendix~\ref{app:sensitivity} examines sensitivity to the per-query budget. We regenerate each query from the same initial latent used by the full-data model. For each metric, we compute the AUC of each run's query--regeneration similarities against its equally sized random-removal control, and report the mean and standard error of the resulting AUCs over the six runs.

\begin{table}[t]
\centering
\caption{Effective removal size, averaged over six seeds, with the fraction of the training set shown in parentheses. Each method selects the top $2\%$ of samples for each of 20 queries; overlap among these per-query sets reduces the union below the nominal total of $40\%$. The random control has the least overlap and therefore removes the most distinct samples.}
\label{tab:removal}
\vspace{1.5mm}
\small
\begin{tabular}{lccc}
\toprule
Method & CIFAR-10 & ArtBench-10 & MS-COCO \\
\midrule
Random           & 16{,}636 (33.3\%) & 16{,}589 (33.2\%) & 39{,}332 (33.3\%) \\
\midrule
CLIP             & 13{,}025 (26.1\%) & 10{,}349 (20.7\%) & 34{,}384 (29.1\%) \\
DINO             & 13{,}728 (27.5\%) & 11{,}857 (23.8\%) & 35{,}952 (30.4\%) \\
D-TRAK           & 15{,}586 (31.2\%) & 13{,}874 (27.8\%) & 35{,}154 (29.7\%) \\
DAS              & 13{,}758 (27.5\%) & 14{,}262 (28.6\%) & 38{,}006 (32.1\%) \\
MUCS             & 15{,}448 (30.9\%) & 14{,}093 (28.2\%) & 34{,}096 (28.8\%) \\
\teacher{} (ours)   & 15{,}419 (30.8\%) & 14{,}231 (28.5\%) & 37{,}467 (31.7\%) \\
\student{} (ours) & 15{,}886 (31.8\%) & 14{,}734 (29.5\%) & 36{,}942 (31.2\%) \\
\bottomrule
\end{tabular}
\end{table}

\paragraph{Method settings.}
Table~\ref{tab:attrib-settings} lists the complete method configurations. \teacher{} uses $M=32$ timesteps from the Karras schedule on every dataset, matching the number of sampling steps. Figure~\ref{fig:efficiency} additionally evaluates $M\in\{1,32,100\}$. We prenormalize each sample's factors and invert the K-FAC factors with relative damping $\alpha=0.1$, setting the damping parameter for each factor to $0.1$ times its mean eigenvalue. For the architecture of \student{} (Eq.~\ref{eq:embeddings}), $\mathrm{AttnPool}$ is a four-head attention layer with a single learned query token, which pools a token sequence into one 768-dimensional vector, and $\mathrm{MLP}$ is a two-layer perceptron ($768 \to 512 \to 768$, SiLU activation). Together they hold $3.2$M trainable parameters ($2.4$M and $0.8$M, respectively), which is only around $3\%$ of the ${\sim}100$M-parameter backbone. The same architecture is used unchanged on all three datasets. We train \student{} for 50 epochs using LambdaRank distillation and a cosine-similarity head. The similarity scale is parameterized as $\alpha=e^{\beta}$, with $\beta$ initialized to $\log 32$, and optimized using the base learning rate of $5\times10^{-5}$. At inference, \student{} also averages scores over $M$ aligned draws. Table~\ref{tab:main} uses $M=32$, while Fig.~\ref{fig:efficiency} evaluates $M\in\{1,32,100\}$; variation across these settings is smaller than the variation across seeds. Baseline hyperparameters follow their official implementations. For DAS, we average projected gradients across timesteps after normalizing them independently at each timestep, use a projection dimension of 16{,}384, and choose regularization within the range evaluated by its authors.

\paragraph{From training pairs to generated queries.}
\student{} is distilled exclusively on within-batch pairs of \emph{training} samples, whereas attribution queries are \emph{generated} samples, introducing a potential train--query distribution shift. Three observations suggest that this shift does not substantially impair performance in our experiments. First, all counterfactual evaluations already test \student{} under this shift: the queries in Table~\ref{tab:main} and Fig.~\ref{fig:efficiency} are generated samples, on which \student{} achieves attribution quality close to that of its teacher. Second, we directly evaluate student--teacher agreement on generated queries using the deployed inference protocol ($M{=}32$ aligned draws, rankings over all training samples, 20 queries, and three runs per dataset). The student's rankings achieve Spearman correlations of $0.87$, $0.89$, and $0.86$ with the teacher's on CIFAR-10, ArtBench-10, and MS-COCO, respectively, demonstrating that the learned embeddings transfer to generated queries. Third, embeddings are computed from \emph{noised} states, where forward noising can reduce differences between the training and generated distributions. This provides a potential mechanism for the observed transfer from training to generation distribution.

\section{Additional Results}
\label{app:additional-results}

\subsection{Qualitative Analysis of Attribution Rankings}

\label{app:qual-topbottom}

\begin{figure}[p]

\centering

\includegraphics[width=\linewidth]{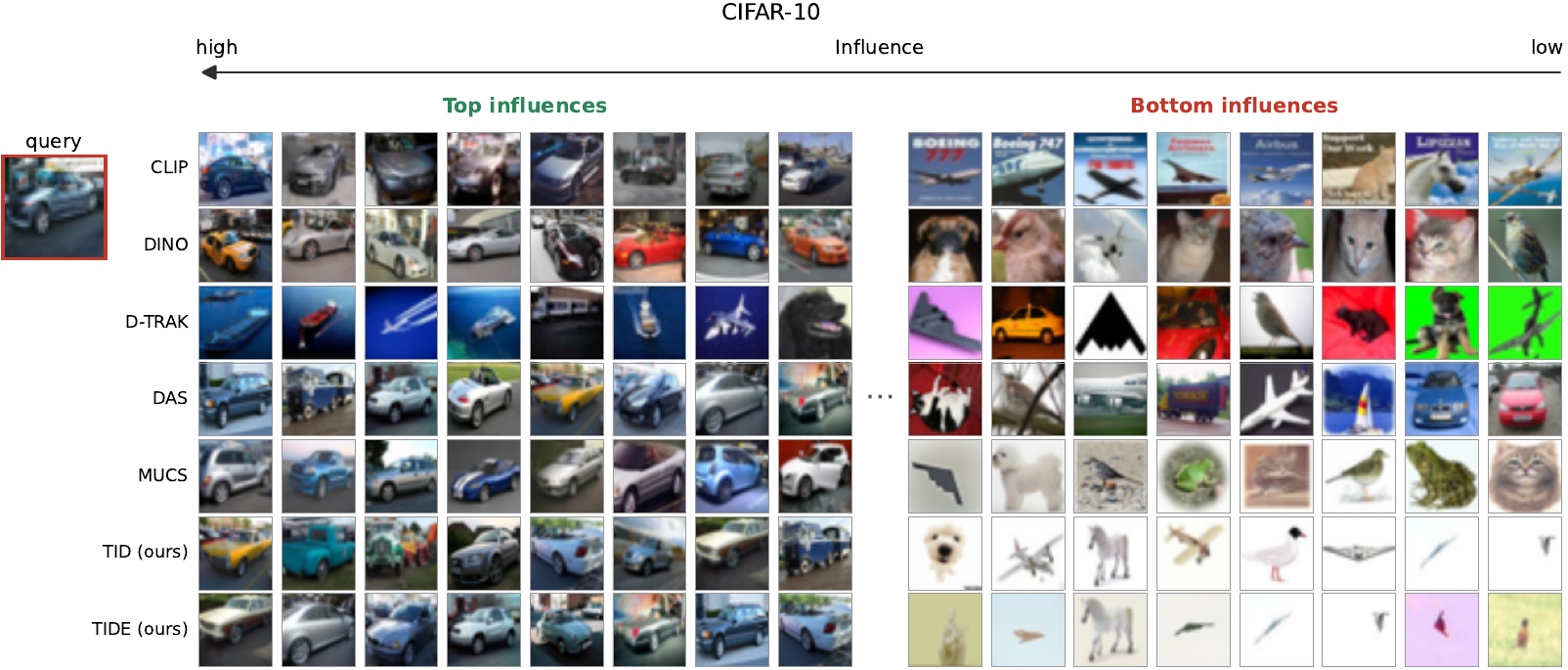}

\caption{Qualitative rankings on CIFAR-10. Each row corresponds to a method in Table~\ref{tab:main} and shows the query, the eight most influential training samples, and the eight least influential samples. The arrow indicates decreasing attributed influence.}

\label{fig:qual-tb-cifar}

\end{figure}

\begin{figure}[p]

\centering

\includegraphics[width=\linewidth]{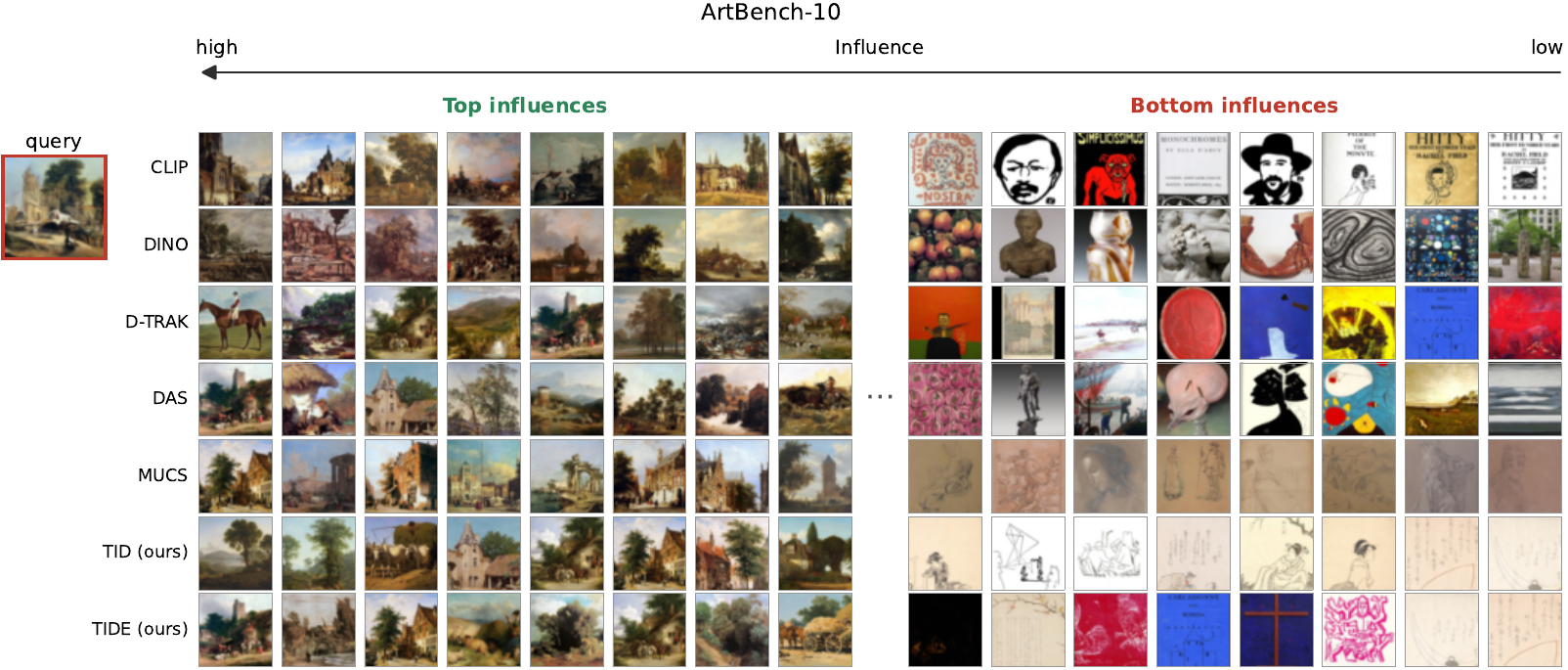}

\caption{Qualitative rankings on ArtBench-10, following the format of Fig.~\ref{fig:qual-tb-cifar}. For this densely painted query, \teacher{} and \student{} place sparse line drawings at the bottom of their rankings.}

\label{fig:qual-tb-artbench}

\end{figure}

\begin{figure}[p]

\centering

\includegraphics[width=\linewidth]{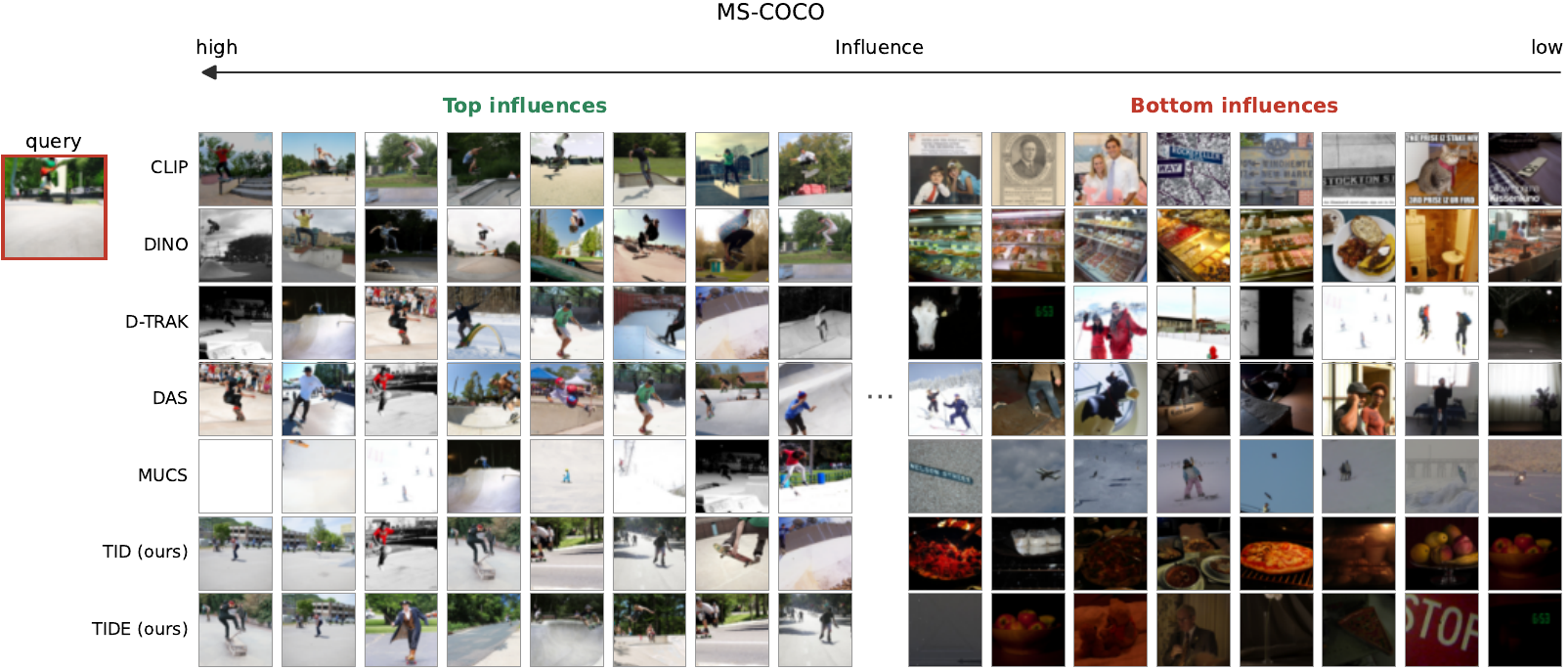}

\caption{Qualitative rankings on MS-COCO, following the format of Fig.~\ref{fig:qual-tb-cifar}. All methods retrieve skateboarding scenes at the top of the ranking, while their lowest-ranked samples differ.}

\label{fig:qual-tb-coco}

\end{figure}

The counterfactual AUC measures the effect of removing selected samples, but does not reveal \emph{which} samples each method selects. Figures~\ref{fig:qual-tb-cifar}--\ref{fig:qual-tb-coco} visualize the rankings for one generated query from each dataset. Each row corresponds to a method in Table~\ref{tab:main} and shows, from left to right, the eight highest-ranked and eight lowest-ranked training samples (we omit the random baseline for clarity). All examples are taken directly from the deployed attribution outputs for the same query; no method is recomputed or re-tuned for this visualization.

The methods differ relatively little at the top of their rankings. For the MS-COCO skateboarding query, all methods retrieve skateboarding scenes. For the CIFAR-10 automobile query, all methods except D-TRAK primarily retrieve automobiles, whereas D-TRAK also selects watercraft, aircraft, and a dog. Because the model-agnostic encoders rank samples using visual similarity alone yet still produce semantically plausible top-ranked examples, agreement at the head of the ranking is not by itself evidence of faithful attribution. This observation is consistent with the relatively weak counterfactual performance of these encoders in Table~\ref{tab:main}.

The tails of the rankings reveal clearer differences. The encoder baselines tend to select semantically dissimilar examples: CLIP retrieves text-heavy posters, while DINO retrieves animals for the automobile query and food displays for the skateboarding query. In contrast, the lowest-ranked samples under \teacher{} and \student{} tend to have weak visual signal, including sparse line drawings on ArtBench-10, isolated objects on white backgrounds on CIFAR-10, and dim nighttime scenes on MS-COCO. This pattern is compatible with a gradient-geometric notion of influence, under which samples can be weakly related to the query even when they are not semantic opposites. DAS exhibits a different pattern on CIFAR-10: its lowest-ranked examples still include cars, boats, and aircraft, suggesting that the two ends of its ranking are not separated by a simple semantic axis.

These single-query examples are intended only to illustrate the content of the rankings. They are consistent with the quantitative results but do not independently establish them.

\subsection{Sensitivity to the Removal Budget}
\label{app:sensitivity}

Our main evaluation removes the top $2\%$ of training samples for each query. To assess the sensitivity to this choice, Table~\ref{tab:removal-sweep} evaluates \student{} on CIFAR-10 using per-query removal budgets of 1, 2, 3, and 5\%. We see that performance improves consistently as the budget increases, indicating that the result is not specific to the default operating point. We also observe a possible trend towards plateau as the budget crosses 3\%, which could suggest that, for CIFAR-10, influence does not propagate significantly over such percentage of training items.

\begin{table}[t]
\centering
\caption{Average counterfactual AUC of \student{} on CIFAR-10 across per-query removal budgets. Performance increases consistently with the budget as larger removals induce more detectable changes in the retrained model.}
\label{tab:removal-sweep}
\vspace{1.5mm}
\begin{tabular}{lcccc}
\toprule
& $1\%$ & $2\%$ & $3\%$ & $5\%$ \\
\midrule
\student{} (ours) & 0.784 & 0.858 & 0.910 & 0.938 \\
\bottomrule
\end{tabular}
\end{table}

\subsection{Relative Magnitude of the Removal Effect}
\label{app:ratio}

In addition to AUC, \citet{serra2026training} report the \emph{similarity ratio}: the percentage change in similarity between each query and its regeneration after removing the attributed samples. Whereas AUC measures how reliably a method can be distinguished from the random-removal control, the ratio measures the magnitude of the induced change. Table~\ref{tab:ratio} reports this quantity for each metric, together with their average, using the same seeds as in Table~\ref{tab:main} (to homogenize results across metrics, in Table~\ref{tab:ratio} we use $1-\text{LPIPS}$, mimicking a similarity metric). More negative values indicate larger removal-induced drift; the random-removal control yields approximately $-$2 to $-$3\%.

\begin{table}[t]
\centering
\caption{Relative change in query--regeneration similarity (\%) after removal, reported as the mean $\pm$ standard error across seeds. More negative values indicate larger induced drift. The random-removal control yields approximately $-$2 to $-$3\%. Method categories follow Table~\ref{tab:main}; the strongest and second-strongest results in each column are shown in \textbf{bold} and \underline{underlined}, respectively.}
\label{tab:ratio}
\small
\begin{tabular}{cllccccc}
\toprule
& Approach & Category & SSIM & SSCD & LPIPS & CLIP & $\mu$ \\
\midrule
\multirow{7}{*}{\rotatebox[origin=c]{90}{CIFAR-10}}
& CLIP & MA & $-3.9{\scriptstyle\pm0.6}$ & $-6.0{\scriptstyle\pm1.1}$ & $-5.1{\scriptstyle\pm0.7}$ & $-7.5{\scriptstyle\pm0.6}$ & $-5.6{\scriptstyle\pm0.7}$ \\
& DINO & MA & $-4.5{\scriptstyle\pm0.7}$ & $-7.5{\scriptstyle\pm0.8}$ & $-6.2{\scriptstyle\pm0.6}$ & $-8.0{\scriptstyle\pm0.9}$ & $-6.5{\scriptstyle\pm0.7}$ \\
& D-TRAK & G & $-6.6{\scriptstyle\pm0.8}$ & $-8.6{\scriptstyle\pm1.1}$ & $-8.2{\scriptstyle\pm0.8}$ & $-7.1{\scriptstyle\pm1.5}$ & $-7.6{\scriptstyle\pm1.0}$ \\
& DAS & G & $-9.2{\scriptstyle\pm1.2}$ & $-14.4{\scriptstyle\pm1.9}$ & $-12.0{\scriptstyle\pm1.4}$ & $-11.8{\scriptstyle\pm1.0}$ & $-11.9{\scriptstyle\pm1.3}$ \\
& MUCS & U & $-12.4{\scriptstyle\pm0.7}$ & $\underline{-19.2}{\scriptstyle\pm0.9}$ & $-16.8{\scriptstyle\pm0.5}$ & $\mathbf{-16.9}{\scriptstyle\pm0.9}$ & $\mathbf{-16.3}{\scriptstyle\pm0.6}$ \\
& \teacher{} (ours) & G & $\underline{-12.7}{\scriptstyle\pm0.8}$ & $-18.1{\scriptstyle\pm1.0}$ & $\underline{-16.9}{\scriptstyle\pm0.5}$ & $\underline{-15.7}{\scriptstyle\pm1.1}$ & $-15.8{\scriptstyle\pm0.8}$ \\
& \student{} (ours) & D & $\mathbf{-13.2}{\scriptstyle\pm0.4}$ & $\mathbf{-19.4}{\scriptstyle\pm0.5}$ & $\mathbf{-17.5}{\scriptstyle\pm0.8}$ & $-15.0{\scriptstyle\pm1.1}$ & $\mathbf{-16.3}{\scriptstyle\pm0.6}$ \\
\midrule
\multirow{7}{*}{\rotatebox[origin=c]{90}{ArtBench-10}}
& CLIP & MA & $-2.4{\scriptstyle\pm1.1}$ & $-5.7{\scriptstyle\pm2.2}$ & $-3.6{\scriptstyle\pm1.8}$ & $-4.8{\scriptstyle\pm1.2}$ & $-4.1{\scriptstyle\pm1.5}$ \\
& DINO & MA & $-2.5{\scriptstyle\pm0.8}$ & $-5.3{\scriptstyle\pm1.8}$ & $-3.5{\scriptstyle\pm0.9}$ & $-3.7{\scriptstyle\pm1.2}$ & $-3.7{\scriptstyle\pm1.1}$ \\
& D-TRAK & G & $-8.7{\scriptstyle\pm1.7}$ & $-14.3{\scriptstyle\pm3.0}$ & $-9.7{\scriptstyle\pm2.4}$ & $-6.6{\scriptstyle\pm1.6}$ & $-9.8{\scriptstyle\pm2.2}$ \\
& DAS & G & $-11.2{\scriptstyle\pm1.5}$ & $-20.7{\scriptstyle\pm1.8}$ & $-13.4{\scriptstyle\pm2.0}$ & $-9.7{\scriptstyle\pm1.2}$ & $-13.7{\scriptstyle\pm1.6}$ \\
& MUCS & U & $\underline{-13.7}{\scriptstyle\pm1.2}$ & $\underline{-23.9}{\scriptstyle\pm1.5}$ & $\underline{-16.3}{\scriptstyle\pm1.6}$ & $\underline{-11.9}{\scriptstyle\pm1.0}$ & $\underline{-16.4}{\scriptstyle\pm1.2}$ \\
& \teacher{} (ours) & G & $\mathbf{-16.4}{\scriptstyle\pm1.3}$ & $\mathbf{-28.6}{\scriptstyle\pm1.9}$ & $\mathbf{-19.2}{\scriptstyle\pm1.6}$ & $\mathbf{-13.1}{\scriptstyle\pm0.8}$ & $\mathbf{-19.3}{\scriptstyle\pm1.3}$ \\
& \student{} (ours) & D & $-10.9{\scriptstyle\pm1.1}$ & $-20.7{\scriptstyle\pm2.1}$ & $-13.6{\scriptstyle\pm1.6}$ & $-9.6{\scriptstyle\pm1.2}$ & $-13.7{\scriptstyle\pm1.4}$ \\
\midrule
\multirow{7}{*}{\rotatebox[origin=c]{90}{MS-COCO}}
& CLIP & MA & $-10.1{\scriptstyle\pm1.1}$ & $-19.5{\scriptstyle\pm3.5}$ & $-11.3{\scriptstyle\pm1.5}$ & $\underline{-11.9}{\scriptstyle\pm0.9}$ & $-13.2{\scriptstyle\pm1.6}$ \\
& DINO & MA & $-11.0{\scriptstyle\pm1.2}$ & $-23.0{\scriptstyle\pm4.0}$ & $-12.7{\scriptstyle\pm1.4}$ & $\mathbf{-12.1}{\scriptstyle\pm0.8}$ & $-14.7{\scriptstyle\pm1.7}$ \\
& D-TRAK & G & $-12.6{\scriptstyle\pm1.1}$ & $-19.7{\scriptstyle\pm3.4}$ & $-13.9{\scriptstyle\pm1.5}$ & $-6.8{\scriptstyle\pm0.7}$ & $-13.2{\scriptstyle\pm1.3}$ \\
& DAS & G & $-12.7{\scriptstyle\pm1.2}$ & $-22.5{\scriptstyle\pm4.0}$ & $-14.7{\scriptstyle\pm1.4}$ & $-8.1{\scriptstyle\pm0.9}$ & $-14.5{\scriptstyle\pm1.7}$ \\
& MUCS & U & $-13.3{\scriptstyle\pm1.0}$ & $-23.6{\scriptstyle\pm2.6}$ & $-15.3{\scriptstyle\pm1.3}$ & $-9.4{\scriptstyle\pm1.2}$ & $-15.4{\scriptstyle\pm1.2}$ \\
& \teacher{} (ours) & G & $\mathbf{-17.6}{\scriptstyle\pm1.3}$ & $\mathbf{-28.8}{\scriptstyle\pm3.9}$ & $\mathbf{-19.8}{\scriptstyle\pm1.1}$ & $-11.3{\scriptstyle\pm1.1}$ & $\mathbf{-19.4}{\scriptstyle\pm1.7}$ \\
& \student{} (ours) & D & $\underline{-14.8}{\scriptstyle\pm1.7}$ & $\underline{-24.6}{\scriptstyle\pm4.1}$ & $\underline{-16.6}{\scriptstyle\pm1.5}$ & $-9.1{\scriptstyle\pm0.7}$ & $\underline{-16.3}{\scriptstyle\pm1.8}$ \\
\bottomrule
\end{tabular}
\end{table}

The ratio results broadly agree with the AUC ranking in Table~\ref{tab:main}. \teacher{} induces the largest average drift on ArtBench-10 and MS-COCO, while \teacher{}, \student{}, and MUCS perform similarly on CIFAR-10. Because the two metrics capture different properties, this agreement serves as a consistency check rather than a redundant evaluation: a method may reliably separate from the random-removal control while inducing only a small absolute change in the regenerated samples.

\subsection{Additional Efficiency Details}

\label{app:efficiency-details}

\paragraph{The complexity split behind Figure~\ref{fig:efficiency}.}
The per-query costs in Figure~\ref{fig:efficiency} are determined by which part of each method's computation is query-independent and small enough to cache. We use the following notation. $N$ is the number of training samples, $Q$ the number of queries, and $M$ the per-sample budget of timesteps or draws. $c_{\mathrm{fwd}}$ and $c_{\mathrm{fb}}$ denote the cost of one forward pass and of one forward--backward pass of the diffusion model $F_\theta$, and $c_{\mathrm{enc}}$ the cost of one pass of an external encoder, $c_{\mathrm{enc,fb}}$ the cost of one forward--backward pass of the encoder. Denote by $E$ the number of distillation epochs, and by $U$ the number of unlearning steps. For the layer-wise factors, $d_l^{\mathrm{in}}$ and $d_l^{\mathrm{out}}$ are the input and output dimensions of layer $l$, and we abbreviate $D_1=\sum_l(d_l^{\mathrm{in}}+d_l^{\mathrm{out}})$, $D_2=\sum_l\big((d_l^{\mathrm{in}})^2+(d_l^{\mathrm{out}})^2\big)$, and $D_3=\sum_l\big((d_l^{\mathrm{in}})^3+(d_l^{\mathrm{out}})^3\big)$, the sizes of the uncompressed factors, of the K-FAC matrices, and the computational cost of their eigendecompositions. Storage is counted in floats. The total model parameters $P=\sum_l\big(d_l^{\mathrm{in}}d_l^{\mathrm{out}}\big)$, and the embedding dimension $d_{\mathrm{emb}}$ and projection dimension $d_{\mathrm{proj}}$ satisfy $d_{\mathrm{emb}} \ll d_{\mathrm{proj}} \ll P$. Writing $C_{\mathrm{total}}$ for the total cost of attributing all $Q$ queries and $C_{\mathrm{cached}}$ for its query-independent part, which is computed once and stored, the warm per-query cost plotted in Figure~\ref{fig:efficiency} is
\begin{equation}
\label{eq:warm-cost}
c_{\mathrm{query}}=\frac{C_{\mathrm{total}}-C_{\mathrm{cached}}}{Q}.
\end{equation}
Table~\ref{tab:complexity} lists both terms per method, separating the flops of the final score products, which are matrix multiplies rather than model evaluations. Below we derive the entries method by method:
\begin{itemize}
\item CLIP and DINO --- The encoder embeds each training sample once, at cost $N c_{\mathrm{enc}}$, giving an $N d_{\mathrm{emb}}$ store. Attributing a query costs one encoder pass and $N\times d_{\mathrm{emb}}$ for inner products.

\item D-TRAK and DAS --- Each training gradient is projected to $d_{\mathrm{proj}}$ dimensions and averaged over $M$ timesteps, so the train side reduces to an $Nd_{\mathrm{proj}}$ store computed once at cost $NM c_{\mathrm{fb}}$ (DAS additionally spends $NM c_{\mathrm{fwd}}$ on its normalization statistics). A query then costs its own $M$ projected gradients, $M c_{\mathrm{fb}}$, and one $N\times d_{\mathrm{proj}}$ vector product. This caching is what random projection buys, at the accuracy cost visible in Table~\ref{tab:main}.

\item TID --- The score in Eq.~\ref{eq:tau-probe} pairs training and query factors from matching draws before averaging, so it cannot be computed from separately pre-averaged factors, and storing the per-draw factors themselves would take $NMD_1$ floats, on the order of a petabyte for CIFAR-10 at $M=32$. Only the K-FAC factors and their eigendecompositions are cacheable, $D_2$ floats or a few GB, computed once at cost $NM c_{\mathrm{fb}}+O(D_3)$. The $NM$ train-side evaluations therefore recur on every pass, and $N$ enters the per-query cost through Eq.~\ref{eq:warm-cost} as $(NM/Q)\,c_{\mathrm{fb}}$, in addition to the query's own $M c_{\mathrm{fb}}$ and an $O(NMD_1)$ score product.

\item TIDE --- Distillation moves all gradient computation into the one-off phase: $O(EN)\,(c_{\mathrm{fb}}+c_{\mathrm{enc,fb}})$ to train the embedder, then $NM (c_{\mathrm{fwd}}+c_{\mathrm{enc}})$ to embed the training set into an $N M d_{\mathrm{emb}}$ store. A query costs $M (c_{\mathrm{fwd}}+c_{\mathrm{enc}})$ and one $N\times M\times d_{\mathrm{emb}}$ vector product, the same warm shape as the encoder baselines.

\item \emph{MUCS} --- No computation is query-independent: each query fine-tunes the model for $U$ steps, $U c_{\mathrm{fb}}$, and then sweeps the training set with $NM$ forward passes, $NM c_{\mathrm{fwd}}$, to measure loss changes, so both $U$ and $NM$ appear in the per-query cost and nothing is cached.
\end{itemize}

\begin{table}[t]
\centering
\caption{Computation and storage by phase. ``One-off'' is computed once on the training set and cached; ``per query'' is Eq.~\ref{eq:warm-cost}. $c_{\mathrm{fwd}}$, $c_{\mathrm{fb}}$, and $c_{\mathrm{enc}}$ are the costs of one forward pass, one forward--backward pass of $F_\theta$, and one external-encoder pass; $D_1$, $D_2$ are the sizes of the uncompressed factors, the K-FAC matrices, and $D_3$ is the computational cost of their eigendecompositions. Cache sizes are in floats; score flops are the matrix multiplies of the final scoring step. We omit the big $O$ notation from the whole table.}
\label{tab:complexity}
\vspace{1.5mm}
\small
\setlength{\tabcolsep}{5pt}
\begin{tabular}{lllll}
\toprule
Method & One-off compute & Cache & Per-query compute & Score flops \\
\midrule
CLIP / DINO & $N c_{\mathrm{enc}}$ & $Nd_{\mathrm{emb}}$ & $c_{\mathrm{enc}}$ & $Nd_{\mathrm{emb}}$ \\
D-TRAK & $NM c_{\mathrm{fb}}$ & $Nd_{\mathrm{proj}}$ & $M c_{\mathrm{fb}}$ & $Nd_{\mathrm{proj}}$ \\
DAS & $NM c_{\mathrm{fwd}} + NM c_{\mathrm{fb}}$ & $Nd_{\mathrm{proj}}$ & $M c_{\mathrm{fb}}$ & $Nd_{\mathrm{proj}}$ \\
\teacher & $NM c_{\mathrm{fb}} + D_3$ & $D_2$ & $\tfrac{NM}{Q} c_{\mathrm{fb}} + M c_{\mathrm{fb}}$ & $NMD_1$ \\
\student & $EN\, (c_{\mathrm{fb}}+c_{\mathrm{enc,fb}}) + NM (c_{\mathrm{fwd}}+c_{\mathrm{enc}})$ & $NMd_{\mathrm{emb}}$ & $M (c_{\mathrm{fwd}}+c_{\mathrm{enc}})$ & $NMd_{\mathrm{emb}}$ \\
MUCS & --- & --- & $U c_{\mathrm{fb}} + NM c_{\mathrm{fwd}}$ & $NM$ \\
\bottomrule
\end{tabular}
\end{table}

\paragraph{Training-side (one-off) costs.}
The warm-cache per-query costs in Figure~\ref{fig:efficiency} exclude the computation performed once on the training set and subsequently reused. Table~\ref{tab:oneoff} below reports these one-off costs and the resulting cache sizes for all three datasets on a single H100. For cacheable methods, preprocessing takes less than five minutes for the model-agnostic encoders, 3.8--10.0\,h for \student{}, and 12.3--38.1\,h for DAS. The \student{} cost is dominated by distillation; constructing the embedding bank itself requires only minutes. After preprocessing, these methods incur only the per-query costs reported in Figure~\ref{fig:efficiency}.

For \teacher{}, factor estimation and eigendecomposition are cacheable, but the training gradients required by each attribution pass are not feasible to store at full scale and must therefore be recomputed (Sec.~\ref{sec:res-efficiency}). MUCS performs no training-side preprocessing and incurs its full computation for every query. The one-off cost also grows faster than the number of training samples alone would predict, because higher-resolution inputs raise the per-sample cost. DAS evaluates $M=100$ timesteps on CIFAR-10 and $M=50$ on the two larger datasets, so normalizing its one-off cost by $|\mathcal{Z}|\cdot M$ isolates the cost of a single image--timestep, which is 13.0, 17.7, and 23.2\,ms on CIFAR-10, ArtBench-10, and MS-COCO, respectively. Scaling the CIFAR-10 measurement by dataset size and timestep count alone would therefore predict 21.4\,h on MS-COCO, well below the measured 38.1\,h; the shortfall is the higher per-node cost of 64$\times$64 inputs with CLIP conditioning.

\begin{table}[t]
\centering
\caption{One-off training-side cost and resulting cache size on a single H100. Encoder and embedding-bank sizes are computed as $N_{\text{train}}d\times4$ bytes; all other cache sizes are measured on disk. The reported wall-clock times are direct measurements from a dedicated timing seed. ``n/a'' indicates that MUCS performs no training-side preprocessing.}
\label{tab:oneoff}
\vspace{1.5mm}
\setlength{\tabcolsep}{6pt}
\resizebox{1\columnwidth}{!}{%
\begin{tabular}{llccc}
\toprule
& & CIFAR-10 & ArtBench-10 & MS-COCO \\
Method & One-off computation & \multicolumn{3}{c}{wall-clock / cache size} \\
\midrule
CLIP   & encode training set
       & 98\,s / 0.10\,GB & 112\,s / 0.10\,GB & 277\,s / 0.24\,GB \\
DINO   & encode training set
       & 18\,s / 0.15\,GB & 18\,s / 0.15\,GB & 37\,s / 0.36\,GB \\
DAS    & error + projected-gradient store
       & 18.1\,h / 3.3\,GB & 12.3\,h / 3.3\,GB & 38.1\,h / 7.8\,GB \\
\student{} & distillation + embedding collection
       & 3.8\,h / 0.15\,GB & 4.2\,h / 0.15\,GB & 10.0\,h / 0.36\,GB \\
\midrule
\teacher{} ($M{=}1$)   & factor estimation
       & 0.2\,h / 2.4\,GB & 0.2\,h / 2.4\,GB & 0.5\,h / 2.5\,GB \\
\teacher{} ($M{=}32$)  & factor estimation
       & 1.7\,h / 2.4\,GB & 2.7\,h / 2.4\,GB & 6.4\,h / 2.5\,GB \\
\teacher{} ($M{=}50$)  & factor estimation
       & 2.6\,h / 2.4\,GB & 4.3\,h / 2.4\,GB & 10.0\,h / 2.5\,GB \\
\teacher{} ($M{=}100$) & factor estimation
       & 5.0\,h / 2.4\,GB & 8.2\,h / 2.4\,GB & 20.1\,h / 2.5\,GB \\
\midrule
MUCS   & none & n/a & n/a & n/a \\
\bottomrule
\end{tabular} }
\end{table}